\documentclass{article}

\usepackage[T1]{fontenc} 
\usepackage[utf8]{inputenc}
\usepackage{amsmath}
\usepackage{amsfonts}
\usepackage{amssymb}
\usepackage{bm}
\usepackage{mathtools}
\usepackage{graphicx}
\usepackage[noend]{algpseudocode}
\usepackage{algorithm}
\usepackage{tikz}
\usepackage{apacite}
\usetikzlibrary{arrows.meta}
\usepackage[left=1in,top=1in,right=1in,bottom=1in,nohead,paperwidth=8.5in, paperheight=11in]{geometry}
\usepackage{natbib}
\usepackage[hidelinks]{hyperref}
\usepackage{booktabs}
\usepackage{longtable}
\usepackage{caption}
\usepackage{setspace}

\newcommand{\ve}[1]{\ensuremath{\boldsymbol{#1}}}
\newcommand{\ma}[1]{\ensuremath{\mathbf{#1}}}
\DeclareMathOperator*{\tr}{tr}

\newcommand{\fracpartial}[2]{\frac{\partial #1}{\partial #2}}
\DeclareMathOperator*{\argmin}{argmin}
\DeclareMathOperator*{\median}{median}

\newcommand{\pkg}[1]{\texttt{#1}}
\newcommand{\proglang}[1]{\textsf{#1}}

\allowdisplaybreaks

\algnewcommand{\TRUE}{\textbf{true}}
\algnewcommand{\FALSE}{\textbf{false}}

\makeatletter
\newcommand{\pushright}[1]{\ifmeasuring@#1\else\omit\hfill$\displaystyle#1$\fi\ignorespaces}
\newcommand{\pushleft}[1]{\ifmeasuring@#1\else\omit$\displaystyle#1$\hfill\fi\ignorespaces}
\makeatother

\def\spacingset#1{\renewcommand{\baselinestretch}%
{#1}\small\normalsize} \spacingset{1}

\title{\bf Clusterpath Gaussian Graphical Modeling}
\author{
Daniel J.W. Touw$^{1}$,
Andreas Alfons$^{1}$, Patrick J.F. Groenen$^{1}$, and Ines Wilms$^{2}$ \\
$^{1}$Department of Econometrics, Erasmus University Rotterdam \\
$^{2}$Department of Quantitative Economics, Maastricht University
}
\date{}

\begin{document}

\maketitle

\begin{abstract}
\noindent
Graphical models serve as effective tools for visualizing conditional dependencies between variables.
However, as the number of variables grows, interpretation becomes increasingly difficult, and estimation uncertainty increases due to the large number of parameters relative to the number of observations.
To address these challenges, we introduce the \textit{Clusterpath estimator of the Gaussian Graphical Model} (CGGM) that encourages variable clustering in the graphical model in a data-driven way.
Through the use of an aggregation penalty, we group variables together, which in turn results in a block-structured precision matrix whose block structure remains preserved in the covariance matrix.
The CGGM estimator is formulated as the solution to a convex optimization problem, making it easy to incorporate other popular penalization schemes which we illustrate through the combination of an aggregation and sparsity penalty.
We present a computationally efficient implementation of the CGGM estimator by using a
cyclic block coordinate descent algorithm.
In simulations, we show that CGGM not only matches, but oftentimes outperforms  other state-of-the-art methods for variable clustering in graphical models.
We also demonstrate CGGM's practical advantages and versatility on a diverse collection of empirical applications.

\bigskip

\end{abstract}

\section{Introduction}
Gaussian graphical models (GGM) are popular tools for summarizing conditional dependencies among $p$ variables.
A GGM is a conditional dependency network where one refers to the variables as the \textit{nodes} and the \textit{edges} represent the conditional dependency relations among each pair of variables.
Estimating GGMs is statistically challenging when the number of parameters ($p(p+1)/2$) is large relative to the sample size ($n$), leading to large estimation variability. Yet, such settings arise across many, diverse fields which has led to a flourishing area on regularized GGM estimation  (e.g., \citealp{meinshausen2006high, peng2009partial, cai2011constrained}).
While much of the existing literature focuses on reducing the estimation variability through edge sparsity, our approach branches into a different direction by using node-clustering. We integrate convex clustering into the GGM framework to reduce estimation variability through estimation pooling  for similar variables.

Edge sparsity through, for instance, $\ell_1$-regularization  has long formed the predominant choice to reduce dimensionality in GGMs (e.g., \citealp{yuan2007graphical_model, yuan2008efficient, friedman2008glasso, rothman2008covar_estimation}).
The idea is to sparsely estimate the precision matrix (i.e., the inverse of the covariance matrix)---the primary object of interest in GGMs---since  conditional independencies between variable-pairs can be directly obtained from its sparsity pattern, or equivalently the absence of edges in the estimated GGM, offering an interpretability advantage.
Several recent studies, however, point to important drawbacks when solely relying on edge sparsity as simplifying structure; namely
weak detection capabilities in large-scale networks \citep{eisenach2020high}, interpretability issues for densely estimated GGMs with many variables \citep{grechkin2015pathway}, or inability to capture real world network structures such as hub nodes \citep{tarzanagh2018estimation} or block-structured graphs \citep{colombi2024learning}.

When GGMs face such challenges, one is often not interested in estimating conditional dependencies among the many observed variables but instead among a smaller number of (unobserved) \textit{clusters} (also called communities) of original variables that share the same behavior.
For instance, biologists estimate large-scale gene regulatory networks and  cluster  genes into pathways to unravel dependencies among them (e.g., \citealp{mestres2018selection, shan2020joint}),
neuroscientists analyzing fMRI data routinely cluster voxels into regions of interest to learn interaction networks of brain activation (e.g., \citealp{pircalabelu2020community}),
financial analysts cluster company stocks into industry sectors to study how shocks can spread over the market by contagion (e.g., \citealp{wilms2021tag_lasso}).
Variable clustering in GGMs not only offers simple, interpretable dependency networks but may also boost the dependency signals \citep{eisenach2020high}.

To estimate GGMs with clustered variables, a recent yet growing interest arose in
\textit{node-based} dimensionality reduction.
While edge sparsity simplifies the graph structure by removing (noisy) edges, variable clustering achieves a simpler graph by aggregating similar nodes. Note that reducing the number of nodes through aggregation automatically reduces the number of edges in the graph as well.
Initial proposals assume the clusters to be known \textit{a~priori} and incorporate this information in a regularization framework \citep[e.g.,][]{grechkin2015pathway, millington2019quantifying}.
Nevertheless, domain-knowledge may not always be available to impose a grouping, calling for unsupervised clustering procedures.
Subsequent works learn the node clustering by decoupling the clustering task from the estimation of the GGM (e.g., \citealp{Ambroise2009InferringSG, tan2015cluster, eisenach2020high, brownlees2022community, shi2024simultaneous}).
Such a two-step procedure may however lead to suboptimal results, giving rise to a third stream that considers both tasks jointly, thereby successfully leveraging a block-structure in the precision matrix to achieve node clustering (e.g., \citealp{yao2019clustered, pircalabelu2020community, wilms2021tag_lasso}).
Yet, to the best of our knowledge, the potential to leverage penalty structures popular in the literature on convex clustering (e.g., \citealp{pelckmans2005convex_clustering, hocking2011convex_clustering, lindsten2011convex_clustering, chi2017cvxbiclust} or, \citealp{weylandt2020dynamic, chakraborty2023biconvex} for more recent advances) to combine node clustering jointly with the estimation of the GGM is left largely unexplored, with \cite{yao2019clustered} being the exception.

We fill this gap by developing a novel regularizer, called the \textit{Clusterpath estimator of the Gaussian Graphical Model} (CGGM), to estimate GGMs that are node-clustered.
Specifically, the cluster structure (i.e.\ the number of clusters and their composition) is identified jointly with the estimation of the parameters.
To this end, we propose a novel penalty on the distances between variables in the precision matrix and embed this in a convex optimization framework for which we offer a computationally efficient cyclic block coordinate descent algorithm (Section~\ref{sec:method}).
The resulting estimated precision matrix has a block structure in which all variables belonging to the same cluster share the same within- as well as cross-cluster dependencies. A unique property of our approach for clustering the precision matrix is that
its inverse retains the same block structure, a property not shared by other approaches---including those discussed below.
Indeed, popular existing paradigms either induce a simplicity structure on the precision matrix or the covariance matrix, and the induced  structure is typically not maintained when taking the inverse of the object of interest. CGGM connects these paradigms by inducing a block structure in the precision matrix that is shared in its covariance matrix.

Our proposal is most closely related to \cite{yao2019clustered}, \cite{pircalabelu2020community}, and \cite{ wilms2021tag_lasso}.
There are two key distinctions with \cite{yao2019clustered}: The first lies in the distance metric of the aggregation penalty: their approach does not account for the diagonal elements of the precision matrix, allowing the diagonal elements of clustered variables to differ. It is due to this property that the variable clustering is not retained when taking the inverse of the estimated precision matrix.
Second, we show how the aggregation penalty of the CGGM estimator can be easily combined with other popular convex penalties such as a sparsity penalty.
Furthermore, the procedure of \cite{pircalabelu2020community} results in a precision matrix with a blockdiagonal structure rather than a full block structure, whereas the regularizer of \cite{wilms2021tag_lasso} requires side-information on the similarity of variables to guide node clustering.

Through a comprehensive simulation study, we evaluate the performance of CGGM against its closest benchmark methods for which software implementations are publicly available (Section \ref{sec:sim}).
Our results indicate that CGGM frequently surpasses the benchmarks in both estimation accuracy and clustering performance.
While the main focus of our study is on estimating clustered precision matrices to create graphical models, we also demonstrate that CGGM can be easily extended to estimate clustered covariance matrices (Section \ref{sec:estimate_cov}). In fact, when a block structure in the covariance matrix (instead of the precision matrix) is the object of interest, we demonstrate that directly estimating a clustered covariance matrix has practical advantages over inverting a clustered precision matrix estimate, even though the latter also yields a block structure in the covariance matrix.
Finally, we illustrate the effectiveness and versatility of CGGM on three practical applications involving (i) stock market data from the S\&P 100, (ii) OECD well-being indicators, and (iii) survey data on participants' humor styles (Section \ref{sec:applications}).

\section{The Clusterpath Estimator for the GGM} \label{sec:method}
We begin in Section~\ref{subsec:GGM} by discussing GGMs with clustered variables, followed by the introduction of the clusterpath estimator in Section~\ref{subsec:clusterpathestimation}.
Section~\ref{subsec:algorithm} details the cyclic block coordinate descent algorithm used to compute our estimator.

\subsection{Clustered GGMs} \label{subsec:GGM}
Let $\ma{X}$ be an $n \times p$ matrix of $n$ multivariate normal observations each of dimension $p$, with sample mean~$\ma{m}$ and sample covariance matrix $\ma{S}$.
Denoting the population covariance matrix by $\ma{\Sigma}$, our target of estimation is the precision matrix $\ma{\Theta} = \ma{\Sigma}^{-1}$.
Under the assumption of a multivariate normal distribution, the precision matrix can be equivalently expressed in a graph where the nodes represent the $p$ variables and the edge weights are given by the entries in $\ma{\Theta}$ which represent the conditional dependencies among the variables.

Our goal is to estimate the precision matrix $\ma{\Theta}$ (and hence the graph structure of the GGM) by encouraging clustering of the $p$ nodes in the graph.
The $K$ new cluster variables $\xi_1, \ldots, \xi_K$ then represent the average of the variables that belong to that cluster, with $K$ typically much smaller than $p$ to achieve dimensionality reduction  in the parameters defining $\ma{\Theta}$.
The edge weights among the clustered variables represent the conditional dependencies among the $K$ new cluster variables. Clustering of the variables in the graph corresponds to a block structure in the rows and columns of $\ma{\Theta}$, see the so-called \emph{G-block} format introduced in \cite{bunea2020model} and discussed by \cite{wilms2021tag_lasso} in the context of GGMs.
In particular, for a partition $\{G_1, \ldots, G_K\}$ of the variables $\{1, \ldots, p \}$ and corresponding $p \times K$ cluster membership matrix $\bf{U}$ with $u_{jk} = 1$ if variable $j$ belongs to cluster $k$ and zero otherwise, there exists a $K \times K$ symmetric matrix $\ma{R}=(r_{k\ell})_{1 \leq k, \ell \leq K}$, and a $p \times p$ diagonal matrix ${\bf A} = \text{diag}(a_{11}\ma{I}, \ldots, a_{KK}\ma{I})$ such that the precision matrix can be written in $G$-block format given by
\begin{eqnarray}
\ma{\Theta} &=& \bf{U} \bf{R} \bf{U}^\top + {\bf A}  \label{theta-block-form}  \\
&=& \begin{small}
\begin{bmatrix}
    r_{11} \ma{11}^\top & r_{12} \ma{11}^\top & \ldots & r_{1K} \ma{11}^\top \\
    r_{21} \ma{11}^\top & r_{22} \ma{11}^\top & \ldots & r_{2K} \ma{11}^\top \\
    \vdots & \vdots & \ddots & \vdots \\
    r_{K1} \ma{11}^\top & r_{K2} \ma{11}^\top & \ldots & r_{KK} \ma{11}^\top
\end{bmatrix} +
\begin{bmatrix}
    a_{11}\ma{I}  & \ma{0} & \ldots & \ma{0} \\
    \ma{0} & a_{22}\ma{I}  & \ldots & \ma{0} \\
    \vdots & \vdots & \ddots & \vdots \\
    \ma{0} & \ma{0} & \ldots & a_{KK}\ma{I}  \\
\end{bmatrix} \end{small}, \nonumber
\end{eqnarray}
where $\ma{I}$ is the identity matrix of appropriate dimension, similarly for $\ma{1}$ denoting a column-vector of ones and for $\ma{0}$ denoting a matrix of zeros.
The within-cluster conditional variances $r_{kk} + a_{kk}$ and covariances $r_{kk}$ are the same for all $p_k$ variables within cluster $k$. These $p_k$ variables in cluster $k$ also have the same conditional covariance $r_{k\ell}$ with all $p_\ell$ variables belonging to another cluster $\ell$.

A fundamental choice made throughout this paper is that $\ma{\Theta}$ should be positive definite. This holds if $\ma{R} + (\ma{U}^\top \ma{U})^{-1} \ma{A}$ is positive definite and $a_{kk} > 0$, for more details, see Appendix~\ref{app:clusterpathderivatation}.
As a consequence of this choice, the range of partial correlations that can be modeled may be limited in certain cases. For example, when $K=1$ and
$\ma\Theta = r_{11}{\bf 1}{\bf 1}^\top + (1 -  r_{11}){\bf I}$, then $r_{11}$ is limited to the range $-1/(p - 1) <  r_{11} < 1$ when requiring $\ma\Theta$ to be positive definite.
Explicit ranges of allowed partial correlations are, however, difficult to formulate more generally for $K>1$ since this depends on the cluster structure.

In contrast to \cite{bunea2020model, wilms2021tag_lasso}, the block structure is not only reflected in the first part of decomposition \eqref{theta-block-form} but also in the diagonal matrix $ {\bf A}$. This means that we impose equal conditional variances of cluster members.
An important implication of such an assumed block structure is that the clustering structure of $\ma{\Theta}$ is thereby retained when taking the inverse \citep[e.g.,][]{gower1991structured, archakov2022canonical}.
The approaches proposed by \cite{yao2019clustered}, \cite{pircalabelu2020community}, and \cite{wilms2021tag_lasso} do not display this property. Section~\ref{sec:estimate_cov} explores the estimation of block-structured covariance matrices using CGGM through a numerical experiment.

\begin{figure}[t]
\centering
\includegraphics[width=0.82\textwidth]{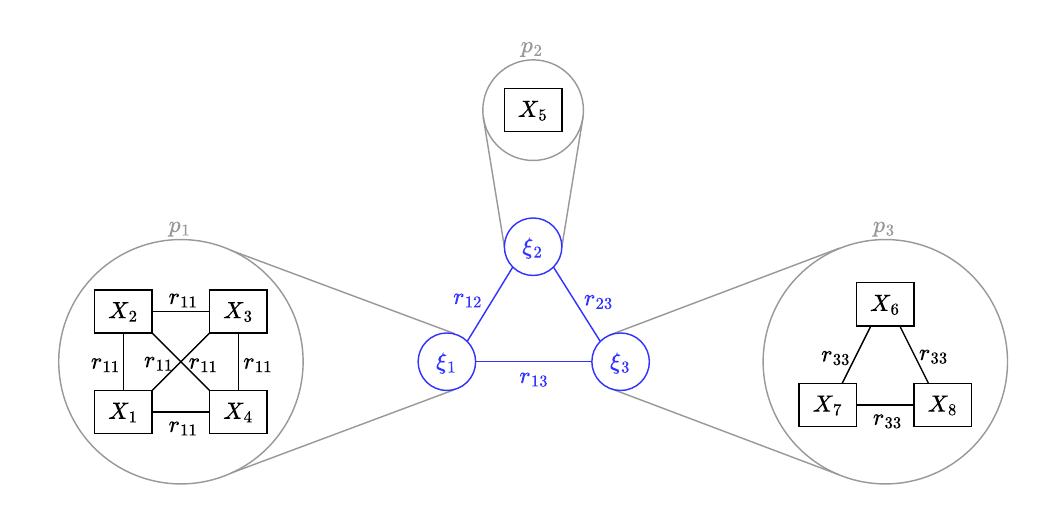}
    \caption{Toy example of a graph representing the clustered precision matrix with $K=3$ clusters constructed from $p=8$  variables. Cluster variable $\xi_1$ is the average of the $p_1 = 4$ variables $X_1, X_2, X_3$, and $X_4$ having within-cluster conditional covariance $r_{11}$. Cluster variable $\xi_2$ is a singleton ($p_2 = 1$) and equal to the original variable $X_5$. Cluster variable $\xi_3$ is the average of the $p_3 = 3$ variables $X_6, X_7$, and $X_8$ having within-cluster conditional covariance $r_{33}$. The three cluster variables have conditional covariances $r_{12}, r_{13},$ and $r_{23}$.}
    \label{fig:graph}
\end{figure}

Next, the block-structured precision matrix can be equivalently visualized through a clustered GGM, see the toy example with $p=8$ variables and $K=3$ clusters in Figure~\ref{fig:graph}.
The block structure in $\ma{\Theta}$ can be interpreted as a clustering in the GGM of variables with identical conditional dependency structure. This cluster structure is a priori, however, unknown and our procedure (Section \ref{subsec:clusterpathestimation}) jointly learns the clustering (including the number of clusters $K$ and their composition) with the estimation of the GGM.
At a coarse level, the clustered GGM visualizes the clustered variables as nodes and displays the conditional dependency structure (edges) among these---the central (blue) part in Figure~\ref{fig:graph}. Since the cluster variables are the averages of its cluster members, this compact visualization implicitly represents the identical conditional dependency structure of all members of a particular cluster with all members of another cluster.
At a more detailed level, the clustered GGM zooms into the members of the newly formed cluster variables---the outer (gray) parts in Figure~\ref{fig:graph}. Cluster members have identical conditional covariances and variances (though the latter is not explicitly visualized in the figure).

Finally, the block structure of the precision matrix in Equation~\eqref{theta-block-form} may display sparsity, where some of the entries in $\ma{\Theta}$ are equal to zero. In particular, the absence of an edge in the clustered GGM between two cluster variables, for instance cluster variables one and two (i.e.\ $r_{12} = 0$), is equivalent to cluster variables one and two being conditionally independent given all other variables, see Proposition 1 in \cite{wilms2021tag_lasso}. In turn, all members of cluster one are then conditionally independent of all members in cluster two.

\subsection{Clusterpath Estimator} \label{subsec:clusterpathestimation}
To estimate a possibly sparse precision matrix with a block structure corresponding to clustered variables, we use a convex penalization method of the form
\begin{align}
    \widehat{\ma{\Theta}} &= \argmin_{\ma{\Theta}} \ L(\ma{\Theta}) \qquad \text{s.t.} \  \ma{\Theta}=\ma{\Theta}^\top, \ma{\Theta} \succ 0, \label{eq:cgm_opt_problem} \\
    L(\ma{\Theta}) &= -\log | \ma{\Theta} | + \tr \ma{S\Theta} + \mathcal{P}(\ma{\Theta}), \label{eq:cgm_loss_main}
\end{align}
where $\log | \cdot|$ denotes the logarithm of the determinant, $\tr(\cdot)$ is  the trace, $\cdot \succ 0$ denotes a positive definite matrix, and $\mathcal{P}(\ma{\Theta})$ is the penalty term. Throughout the paper we take
\begin{equation} \label{eq:penalty-term}
    \mathcal{P}(\ma{\Theta}) = \lambda_c \sum_{j=2}^p\sum_{j'=1}^{j-1} w_{jj'} d_{jj'}(\ma{\Theta}) + \lambda_s \sum_{\substack{j, j' \\ j'\neq j}} z_{jj'} |\theta_{jj'}|,
\end{equation}
with
\begin{equation*}
    d_{jj'}(\ma{\Theta}) = \sqrt{(\theta_{jj} - \theta_{j'j'})^2 + \sum_{\substack{m=1\\ m \notin \{j, j'\}}}^p(\theta_{jm} - \theta_{j'm})^2}.
\end{equation*}
The first part in \eqref{eq:penalty-term} represents the aggregation penalty penalizing differences $d_{jj'}(\ma{\Theta})$ between columns $\bm{\theta}_j= (\theta_{1j},...,\theta_{pj})^\top$ and $\bm{\theta}_{j'}$ of the precision matrix $\ma{\Theta}$. Its role is thus to encourage estimation of a precision matrix with a $G$-block structure; or, put alternatively, a GGM with clustered variables. The second part represents the  sparsity penalty on the unique off-diagonal elements of the precision matrix with sparsity weights $z_{jj'}$, defined in Appendix~\ref{app:Ltheta_reexpressed}. Its role is to encourage estimation of precision matrices with zero off-diagonal entries; or, put alternatively, GGMs with edge-sparsity.
The tuning parameters $\lambda_c$ and $\lambda_s$ control the degree of aggregation and sparsity respectively. The objective function is convex since each of the terms  in \eqref{eq:cgm_loss_main} is convex in $\ma{\Theta}$ and $\ma{\Theta}$ lies in the convex cone of positive definite matrices.

Focusing on the aggregation penalty, the weights $w_{jj'}$, for every unique pair $1 \leq j, j' \leq p$, determine the fundamental attraction between two variables and are specified beforehand based on, for example, domain knowledge or the pairwise distances based on the sample precision matrix $\ma{S}^{-1}$. We further discuss the choice of weights in Section \ref{subsec:algorithm}.
If $d_{jj'}(\bm{\Theta}) = 0$ and $\lambda_c$ is positive, then two vectors ${\bm{\theta}}_j$ and ${\bm{\theta}}_{j'}$ have exactly the same elements while ignoring
 ${\theta}_{jj'}$ and ${\theta}_{j'j}$.
The distance $d_{jj'}(\ma{\Theta})$ disregards the difference between $\theta_{jj'}$ and $\theta_{j'j}$ as these are identical due to the symmetry of $\ma{\Theta}$.
We do also penalize the differences between entries on the diagonal, since the block structure must also be reflected in the diagonal of the precision matrix, see Equation~\eqref{theta-block-form}.
The penalty term $d_{jj'}(\bm{\Theta})$ can thus be interpreted as a group lasso penalty \citep{yuan2006group}, where each pair of variables $j < j'$ forms a group whose elements are the differences between the  corresponding entries in the columns $\bm{\theta}_j$ and $\bm{\theta}_{j'}$ of the precision matrix.
If all respective differences are put to zero, the estimated entries are identical, which in turn effectively blocks columns $j$ and $j'$ in the precision matrix, or equivalently clusters nodes $j$ and $j'$ in the GGM.
The shrinkage of the distances between the clusters pushes the corresponding entries in the precision matrix closer to each other across clusters, thereby introducing bias; we accommodate this through a common `post-selection' refitting step in the regularization literature \citep{BelloniChernozhukov2013}, see Section \ref{subsubsec:refit}.
As $\lambda_c$ increases (and $\lambda_s$ is fixed), estimated GGMs are obtained in which more and more variables are clustered, thereby resulting in a \textit{clusterpath} from the $p$ original nodes (no clustering) until one clustered node (full clustering) in the GGM.

The notion of applying a distance-based penalty to pairs of objects stems from convex clustering \citep{pelckmans2005convex_clustering, hocking2011convex_clustering, lindsten2011convex_clustering}.
In convex clustering, a copy of the data matrix is estimated while penalizing the distances between rows, thereby facilitating the clustering of observations.
Building upon this framework, convex biclustering also clusters variables by adding a penalty on the distances between columns \citep{chi2017cvxbiclust}.
Leveraging this notion for clustering variables via the precision matrix offers insights into the underlying conditional dependencies within the data.

\subsection{Cyclic Block Coordinate Descent Algorithm} \label{subsec:algorithm}

We develop a cyclic block coordinate descent algorithm, tailored to minimizing objective function \eqref{eq:cgm_opt_problem}.\footnote{
One could opt for alternative algorithms such as the  alternating directions method of multipliers (ADMM), provided it is tailored towards solving problem \eqref{eq:cgm_opt_problem}; as for instance \cite{wilms2021tag_lasso} do for their method. However, while all the subproblems of the ADMM by \cite{wilms2021tag_lasso}  are solvable in closed-form, this would not be the case for the subproblem corresponding to our proposed aggregation penalty.
This reason, in combination with the natural link between the block structure in $\boldsymbol\Theta$ and the block structure of the cyclic block coordinate descent algorithm further explains our choice for the latter.
}
We opt for cyclic block coordinate descent since the blocks are naturally formed by the clusters $k=1, \ldots, K$. By exploiting this block structure in $\ma{\Theta}$, our algorithm builds on efficient expressions of the objective function, which allows us to minimize optimization problem \eqref{eq:cgm_opt_problem} in a computationally efficient way.

For ease of exposition, first suppose that the block structure of the precision matrix in Equation~\eqref{theta-block-form} is known; hence the number of clusters $K$ and the cluster membership is known. To permit cyclic block updates, one for each block/cluster $k$, the objective $L(\ma{\Theta})$ needs to be separated into those parts that depend on cluster $k$ and those that do not. To this end, we re-write the precision matrix as
\begin{align*}
    \ma{\Theta}
    &= \left[
    \begin{array}{@{}c|c@{}}
         \ma{\Theta}_{00} & \ma{\Theta}_{0k} \\\hline
         \ma{\Theta}_{0k}^\top & \ma{\Theta}_{kk}
    \end{array}\right]
     = \left[\begin{array}{@{}c|c@{}}
         \ma{\Theta}_{00} & \ma{0} \\\hline
         \ma{0} & \ma{0}
    \end{array}\right] +
    \left[\begin{array}{@{}c|c@{}}
         \ma{0} & \ma{\Theta}_{0k} \\\hline
         \ma{\Theta}_{0k}^\top & \ma{\Theta}_{kk}
    \end{array}\right] =
    \ma{\Theta}_{-k} + \ma{\Theta}_k,
\end{align*}%
where the first equality splits the precision matrix into four blocks: $\ma{\Theta}_{00}$ contains all elements not pertaining to cluster $k$, $\ma{\Theta}_{0k}$ contains all cross-cluster conditional covariances involving cluster $k$, and  $\ma{\Theta}_{kk}$ contains all within-cluster $k$ conditional variances and covariances.
Without loss of generality, the assumption can be made that the variables are arranged to consistently position cluster $k$ as the last cluster. Finally, we denote all parameters in $\ma{\Theta}$ belonging to cluster $k$ as $\ma{\Theta}_k$, thereby making use of single subscript notation. Similarly, $\ma{\Theta}_{-k}$ collects all parameters in $\ma{\Theta}$ not belonging to cluster $k$. The objective function can then be partitioned into four distinct components, as given by
\begin{equation}
\label{eq:loss_per_k}
    L(\ma{\Theta}_k) = L_{\text{det}}(\ma{\Theta}_k) + L_{\text{cov}}(\ma{\Theta}_k)  +  L_{\text{clust}}(\ma{\Theta}_k) + L_{\text{sparse}}(\ma{\Theta}_k) + C,
\end{equation}
where $L_{\text{det}}$, $L_{\text{cov}}$, $L_{\text{clust}}$, and $L_{\text{sparse}}$ represent the log-determinant, trace, and penalty parts (cluster- and sparsity-based) of the objective function that depend on cluster $k$, and finally $C$ is a constant collecting parts independent of cluster $k$. Details of these expressions are in Appendix~\ref{app:clusterpathderivatation}.
The cyclic block coordinate descent algorithm then simply cycles through all blocks $k=1, \ldots, K$, thereby optimizing  $L(\ma{\Theta}_k)$ in the $k$th cluster.
When optimizing $L(\ma{\Theta}_k)$, we do this computationally efficiently by building on the $G$-block format of the precision matrix which re-parametrizes $\ma{\Theta}$  in terms of $\ma{A}$ and $\ma{R}$, see Equation~\eqref{theta-block-form}. The re-parametrization permits more efficient updates when optimizing  $L(\ma{\Theta}_k)$; in particular more efficient expressions of $L(\ma{\Theta}_k)$, and of the gradient and Hessian of \eqref{eq:loss_per_k}, as detailed in Appendix~\ref{app:clusterpathderivatation}.

In practice, the block structure of the precision matrix is not known beforehand; our penalization problem \eqref{eq:cgm_opt_problem} jointly identifies the block structure and estimates the precision matrix $\ma{\Theta}$.
The identification of the block structure, meaning the estimation of the number of clusters $K$ and cluster composition, is done as follows.
Every variable is initialized in its own cluster (i.e.\ $\hat{K}=p$).
For a fixed value of the aggregation penalty $\lambda_c$ and initial estimate at $\widehat{\ma{\Theta}}$, the clusters that are eligible for fusion are determined by computing the distances $d_{jj'}(\widehat{\ma{\Theta}})$ for every variable pair $j, j'$. If this distance reduces to zero,  the two corresponding variables $j$ and $j'$ are eligible for fusion.
We work with a threshold $\epsilon_f$  to assess a fusion in practice (see Section \ref{subsubsec:algorithm}) due to practical limitations in numerical accuracy.
The number of clusters $\hat{K}$ and cluster composition are then updated accordingly. Given this updated block structure, objective function \eqref{eq:loss_per_k} is optimized for every block/cluster $k=1, \ldots, \hat{K}$.

In sum, the proposed cyclic block coordinate descent algorithm involves two key steps. For each cluster $k$, it first assesses whether there is an eligible fusion candidate. In the instances where no suitable candidate is identified, the algorithm proceeds to the second step, wherein the parameters associated with cluster $k$ are updated using Newton's method.
Another reason we choose a cyclic block  coordinate descent algorithm lies with the use of Newton's method since the computation of a Newton descent direction for all parameters in the precision matrix simultaneously would result in a computationally intractable algorithm even for moderately sized data sets.

In Section \ref{subsubsec:algorithm}, we provide details on the two key steps of the optimization algorithm.
We further discuss two important components related to the
penalty terms, namely the weight matrix (Section \ref{subsubsec:weights}) and the tuning parameter (Section \ref{subsubsec:tuningparam}).
Finally, in Section~\ref{subsubsec:refit}, we discuss a refitting step to reduce bias in the CGGM estimate.

\subsubsection{Outline of the Cyclic Block Coordinate Descent Algorithm} \label{subsubsec:algorithm}
We outline the main two steps of the cyclic block coordinate descent algorithm minimizing the objective function in \eqref{eq:loss_per_k} iteratively for cluster $1 \leq k \leq K$ for fixed values of the tuning parameters $\lambda_c$ and $\lambda_s$.
The pseudo-code for the algorithm is presented and discussed in Appendix~\ref{appendix:algorithm}.
The computational complexity of the algorithm is $\mathcal{O}(K^4)$, which is particularly efficient when the number of clusters $K$ is small relative to the number of variables $p$.
Moreover, the cyclic updates facilitate the preservation of the positive definiteness of $\ma{\Theta}$ as long as each update satisfies two straightforward inequalities, detailed in Appendix~\ref{appendix:algorithm}.

{\bf Cluster fusions.} The first step is to verify whether cluster $k$ is close enough to another cluster to warrant their fusion for estimate $\widehat{\ma{\Theta}}$ (in a given iteration of the algorithm).
Let $\mathcal{C}_k$  denote the set of variables belonging to cluster $k$, analogously $\mathcal{C}_\ell$ for cluster $\ell$. Then,
cluster $k$ is fused with cluster $\ell$ if $d_{\mathcal{C}_k \mathcal{C}_\ell}(\widehat{\ma{\Theta}})$, which measures the distance between blocks $\widehat{\ma{\Theta}}_k$ and $\widehat{\ma{\Theta}}_\ell$, is smaller than some user-defined threshold $\varepsilon_f$ representing the minimum required similarity for fusion.
In case of multiple candidates for which this applies, cluster $\ell$ is chosen as $\ell= \argmin_{\ell'}d_{\mathcal{C}_k \mathcal{C}_{\ell'}}(\widehat{\ma{\Theta}})$. If a fusion is performed, it advances to the next cluster. If no eligible fusion candidate is found in the first step, we proceed to the second step where we update the parameter estimates for cluster $k$.

{\bf Parameter estimation.} The second step consists of updating the parameters estimates for cluster $k$ using Newton's method.
We opt for a Newton descent direction as it offers efficient convergence properties. Let $\nabla L(\widehat{\ma{\Theta}}_k)$ denote the gradient and $\nabla^2 L(\widehat{\ma{\Theta}}_k)$ the Hessian of $L(\widehat{\ma{\Theta}}_k)$.
The descent direction is then computed as
    $\bm\delta_k = -\nabla^2 L(\widehat{\ma{\Theta}}_k)^{-1} \nabla L(\widehat{\ma{\Theta}}_k)$, see Appendix~\ref{app:subsec:gradientandhessian} for the derivations.

Finally, we determine the optimal step size $s^*$ given the descent direction using an inexact line search. The reason for augmenting the Newton descent direction with a line search is twofold. First, the minimization of $L(\ma{\Theta}_k)$ is constrained by the restriction that $\ma{\Theta}$ should be positive definite. Hence, a step size should be chosen that ensures the update to adhere this restriction.
In Appendix~\ref{appendix:algorithm}, we derive the conditions under which the optimal step size can be computed that adheres to the positive definiteness restriction.
A second reason to opt for a line search is the fact that the Hessian may not be well-behaved due to the presence of the $\ell_2$-norm in the clusterpath penalty. If, for the current value
$\widehat{\ma{\Theta}}$, at least one of the distances $d_{\mathcal{C}_k \mathcal{C}_\ell}(\widehat{\ma{\Theta}})$  is close to zero, $L(\widehat{\ma{\Theta}}_k)$ is not locally smooth. As a result, the Hessian is not well-behaved and the favorable convergence properties of Newton's method do not hold. A line search for the optimal step size guarantees that the update does not increase the objective function.

\subsubsection{Weight Matrix} \label{subsubsec:weights}
An important element of the clusterpath estimator is the weight matrix $\ma{W}$ that contains information about the preferences of clustering variables $j$ and $j'$ through the weight $w_{jj'}$. A large value for $w_{jj'}$ incentivizes clustering of the corresponding variables, but it does not guarantee this.
In the convex clustering literature, weights are typically based on the squared distances between data points, with performance improving when using $k$-nearest neighbors to assign zero weights to nonneighbor pairs \citep[e.g.,][]{chen2015convex, chi2015ama}.
Following these standard practices, we scale the squared distances by the mean squared distance and take
\begin{equation}
    w_{jj'} = w_{j'j} =
    \left\{
    \begin{array}{@{}ll}
        \exp\left(-\phi \frac{d^2_{jj'}(\ma{S}^{-1})}{\sum_{(m, m') \in \mathcal{E}}d^2_{mm'}(\ma{S}^{-1}) / |\mathcal{E}|}\right)
        & \quad \text{if $(j, j') \in \mathcal{E}$}, \\
        0 & \quad \text{otherwise},
    \end{array}
    \right. \label{eq:weights}
\end{equation}
where $\phi$ is a tuning parameter, $\mathcal{E}$ is the set of variable pairs $(j, j')$ that should be assigned a nonzero weight, and $|\mathcal{E}|$ denotes the number of elements in $\mathcal{E}$.
Note that the choice of $w_{jj'}$ in~\eqref{eq:weights} is only available if $\ma{S}$ is invertible. If $\ma{S}$ is not invertible, one can use a regularized version instead. Throughout the paper, we use $(\ma{S} + \ma{I})^{-1}$ in such cases, but the user may resort to other regularized alternatives such as the graphical lasso.
For the hyperparameter $\phi$, it can be beneficial to choose candidate values such that the nonzero weights are distributed broadly over the interval $(0, 1]$ without large peaks at specific values; we provide practical guidance in the applications discussed in Section \ref{sec:applications}.

A potential issue with relying solely on $k$-nearest neighbors for the construction of $\mathcal{E}$ is the occurrence of disconnected subgroups. To alleviate this issue, we add variable pairs to~$\mathcal{E}$ using the procedure described in \cite{touw2023convex} until all disconnected subgroups are eliminated.
This approach effectively integrates the $k$-nearest neighbors structure with a minimum spanning tree to obtain a weight matrix corresponding to a connected graph.

\subsubsection{Tuning Parameters} \label{subsubsec:tuningparam}
The tuning parameters $\lambda_c$ and $\lambda_s$ are other key ingredients of CGGM as they control the degree of clustering and sparsity respectively.

To set a sequence for $\lambda_c$ (for given values of $\lambda_s$ and $\ma{W}$), we implement an automated procedure that ensures a continuum of solutions for $\ma{\Theta}$ with smooth transition from minimal (where $\hat{K}=p$) to maximal regularization  (where $\hat{K}=1$). Details are provided in Appendix~\ref{app:subsec:alg_details}.
Furthermore, note that for
each subsequent optimization problem with a larger value for $\lambda_c$, we make use of warm starts provided by the solution to the previous problem, see Appendix~\ref{appendix:algorithm}. This warm start includes the clustering information for that solution.

For the sparsity parameter $\lambda_s$, we use a grid of ten values with each step being twice as large as the previous step. The largest value in the grid is based on the smallest value of $\lambda_s$ that puts all off-diagonal elements in $\ma{\Theta}$ to zero for the graphical lasso (i.e.\ for  $\lambda_c=0$). The other values in the grid are set as fractions (from zero to one) of this maximum value.

Finally, to select the tuning parameters, we adopt a cross-validation procedure.
That is, we find the combination of tuning parameters that minimizes the cross-validated likelihood-based score given by
\begin{equation}
\label{eq:cv_score}
    \frac{1}{G} \sum_{g=1}^G  -\log | \widehat{\ma{\Theta}}_{-\mathcal{F}_g} | + \tr \ma{S}_{\mathcal{F}_g}\widehat{\ma{\Theta}}_{-\mathcal{F}_g},
\end{equation}
where $\widehat{\ma{\Theta}}_{-\mathcal{F}_g}$ represents the precision matrix estimated on the sample excluding the observations in the $g$\textsuperscript{th} fold, and $\ma{S}_{\mathcal{F}_g}$ denotes the sample covariance matrix computed solely on the samples within the $g$\textsuperscript{th} fold.

\subsubsection{Refitted CGGM}
\label{subsubsec:refit}
While minimization of the objective function in \eqref{eq:cgm_opt_problem} yields a possibly sparse and/or clustered $\widehat{\ma{\Theta}}$, the CGGM estimate may be biased due to the shrinkage of the distances between different clusters and the shrinkage of the entries in the precision matrix to zero.
Following \cite{wilms2021tag_lasso}, we therefore also consider a refitted version of CGGM.
First, CGGM is used to obtain a clustering and sparsity pattern in $\widehat{\ma{\Theta}}$.
We then re-estimate the  precision matrix $\ma{\Theta}$  constrained by the obtained clustering and sparsity by maximizing the likelihood subject to the clustering and sparsity constraint.
By re-estimating the precision matrix, we aim to capitalize on the clustering- and sparsity-inducing capabilities of CGGM while refining parameter estimates for improved accuracy.

\section{Simulations}
\label{sec:sim}

To investigate the behavior of the proposed method CGGM, we perform an extensive simulation study. We discuss the data generating processes in Section~\ref{sec:sim-design}, the benchmark methods and evaluation criteria in Section~\ref{sec:sim-methods}, and the results in Section~\ref{sec:sim-results}.

\subsection{Simulation Designs}
\label{sec:sim-design}

For most of the simulations, we follow the designs of \citet{wilms2021tag_lasso} together with additional variations. In all settings, we sample data from a multivariate normal distribution with mean zero and covariance $\ma{\Sigma} = \ma{\Theta}^{-1}$, and we repeat the process 100 times.

\begin{figure}[!t]
\begin{center}
\includegraphics[width=0.8\textwidth]{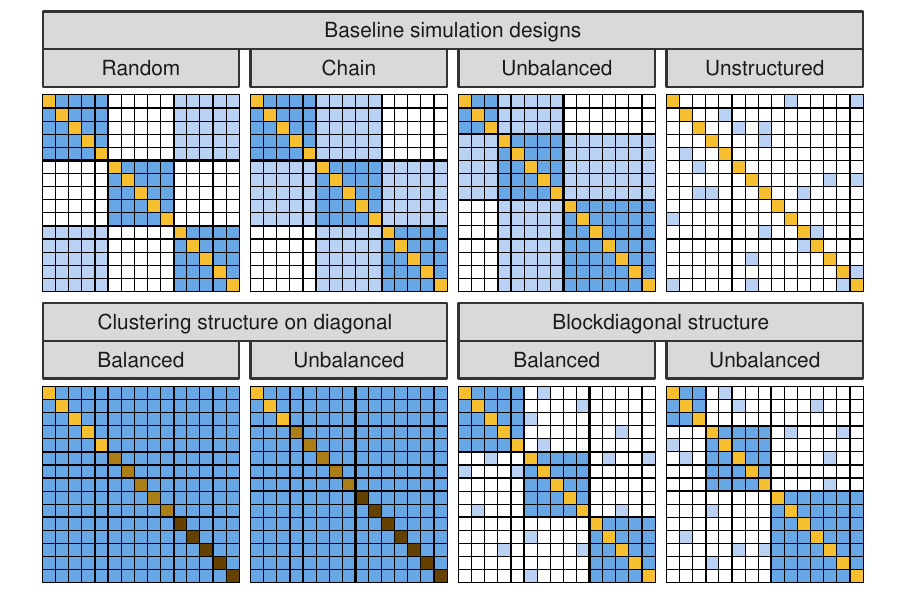} \vspace{-12pt}
\end{center}
\caption{Precision matrices \ma{\Theta} in the baseline simulation designs (top), the designs with clustering structure on the diagonal and with blockdiagonal structure, respectively, using balanced and unbalanced clusters sizes (bottom).
The color shade indicates the magnitude of the elements.
Diagonal elements are on a separate color scale than the off-diagonal ones to highlight their differing roles in CGGM in comparison to the benchmark methods.}
\label{fig:theta_designs}
\end{figure}

{\bf Baseline simulation designs.}
We simulate $n=120$ observations on $p=15$ variables using four different structures for the precision matrix $\ma{\Theta}$ (see the top row of Figure~\ref{fig:theta_designs}):
(i)~\textit{random}: the clusters are of equal size with $p_{1} = p_{2} = p_{3} = 5$, and one pair of clusters is selected at random to be connected via an edge;
(ii)~\emph{chain}: the clusters are of equal size with $p_{1} = p_{2} = p_{3} = 5$, and adjacent clusters are connected via an edge;
(iii)~\emph{unbalanced}: same as the chain design, but the clusters are of unequal size with $p_{1} = 3$, $p_{2} = 5$, and $p_{3} = 7$;
(iv)~\emph{unstructured}: each variable forms its own cluster, and edges between variables are drawn with probability $\pi = 0.1$.
The first three designs thus employ a block structure with $K=3$ clusters, while the fourth  does not exhibit any variable aggregation structure. In all four designs, the diagonal elements of $\ma{\Theta}$ are set to 1, the elements within a cluster of variables to 0.5, and the non-zero elements between clusters to 0.25.

{\bf Increasing the number of variables.}
We focus on the chain design as in \citet{wilms2021tag_lasso} and vary the number of variables $p \in \{15, 30, 60, 120\}$. We set the corresponding number of observations $n \in \{120, 240, 480, 960\}$ such that the ratio $n/p$ is constant. The number of clusters is kept fixed at $K = 3$.

{\bf Increasing the number of clusters.}  We again focus on the chain design and vary the number of clusters $K \in \{3, 5, 6, 10\}$ while keeping the number of observations and variables fixed at $n = 240$ and $p = 30$, respectively, as in \citet{wilms2021tag_lasso}.

{\bf Approximate block structure.}
We take the four baseline designs and modify the structure in $\ma{\Theta}$ by uniformly drawing elements within a cluster of variables from the interval $[0.4, 0.6]$ and non-zero elements between clusters from the interval $[0.2, 0.3]$. The assumptions of CGGM are not (fully) met in the resulting designs, as the clusters do not correspond to blocks of equal values in $\ma{\Theta}$ but blocks of approximately similar values.

{\bf Clustering structure on diagonal.} Using the same dimensions and number of clusters as in the baseline designs, all off-diagonal elements in $\ma{\Theta}$ are set to 0.5. The diagonal elements are set to 1 for the variables in the first cluster, to 2 for the variables in the second cluster, and to 3 for the variables in the third cluster. We investigate both a \emph{balanced} design with $p_{1} = p_{2} = p_{3} = 5$, and an \emph{unbalanced} design with $p_{1} = 3$, $p_{2} = 5$, and $p_{3} = 7$ (see left two panels in the bottom row of Figure~\ref{fig:theta_designs}).

{\bf Blockdiagonal structure.} In these designs, the precision matrix $\ma{\Theta}$ does not exhibit a variable clustering structure, but a blockdiagonal structure inspired by the simulation design of \citet{pircalabelu2020community}. We use the same dimensions as in the baseline design and $K=3$ blocks on the diagonal. Outside of the diagonal blocks, edges between variables are drawn with probability $\pi = 0.1$. The diagonal values of $\ma{\Theta}$ are set to 1, the off-diagonal elements in the diagonal blocks are set to 0.5, and non-zero elements outside the diagonal blocks are set to 0.25. If such a randomly drawn precision matrix is not positive semi-definite, we repeat this process until a positive semi-definite precision matrix is obtained. We consider both a \emph{balanced} design with $p_{1} = p_{2} = p_{3} = 5$, and an \emph{unbalanced} design with $p_{1} = 3$, $p_{2} = 5$, and $p_{3} = 7$ (see right two panels in bottom row of Figure~\ref{fig:theta_designs}).

\subsection{Methods and Evaluation Criteria}
\label{sec:sim-methods}

We apply CGGM with and without the parameter re-estimation step described in Section~\ref{subsubsec:refit}, which we refer to as \emph{CGGM-raw} and \emph{CGGM-refit}. We thereby set the convergence tolerance $\varepsilon_c = 10^{-7}$ and the maximum number of iterations to $t_\text{max} = 5000$. For comparison, we include the tree-aggregated graphical lasso (\emph{TAGL}) of \citet{wilms2021tag_lasso}, the community-based group graphical lasso (\emph{ComGGL}) of \citet{pircalabelu2020community}, the graphical lasso (\emph{GL}) \citep{friedman2008glasso}, and the inverse of the sample covariance matrix ($\ma{S}^{-1}$).

TAGL performs node aggregation based on side-information in the form of a tree-based variable hierarchy, which needs to be specified a priori. We generate an \emph{ideal} and a \emph{realistic} tree hierarchy as described in \citet{wilms2021tag_lasso} (see Figure~5 in their paper for an illustration). As both trees contain the true variable clustering, we also generate a \emph{misspecified} tree  in the same manner as the realistic tree, except that with probability 0.1, each variable is incorrectly assigned to the subsequent cluster (or the first cluster if the variable in fact belongs to the last cluster).

ComGGL does not perform node aggregation but community detection in the form of a blockdiagonal structure.
Since ComGGL merely encourages block-diagonality in the precision matrix, the entries of the precision matrix estimate corresponding to features belonging to the same community may still vary.
Hence, ComGGL does not induce block-structured precision matrix estimates but serves mostly as a relevant benchmark for the simulation designs with approximate block structure and blockdiagonal structure.

Concerning tuning parameter selection, we apply 3-fold cross-validation. For the weight matrix in CGGM, we use candidate values $k \in \{1, 3, 5\}$ for the number of neighbors and we set $\phi=1$ to keep the computational burden low. We verified that for this choice of $\phi$, the weights showed sufficient variation over the interval $(0, 1]$ to stimulate variable clustering.
Moreover, we select candidate value for the regularization parameters $\lambda_c$ and $\lambda_s$ as described in Section~\ref{subsubsec:tuningparam}.
For TAGL, we first perform a binary search for the smallest value of the aggregation parameter that aggregates all variables into one cluster (for each of the tree hierarchies) while keeping the sparsity parameter at~0, and we determine the smallest value of the sparsity parameter that removes all edges while keeping the aggregation parameter at 0 (corresponding to the graphical lasso).
Ten candidate values for the aggregation and sparsity parameters are then obtained as fractions of those aforementioned values, where the fractions vary from 0 to 1, with each step being twice as large as the previous step. For ComGGL, we similarly set ten candidate values for the grouping and sparsity parameters as fractions of maximum values. While we fix the maximum value of the grouping parameter at 1, we again set the maximum value of the sparsity parameter as the smallest value that removes all edges in the graphical lasso. As in \citet{pircalabelu2020community}, we set the balancing parameter to 1. For the graphical lasso, we choose ten candidate values for the sparsity parameter in the same manner.

We evaluate the methods in terms of their estimation accuracy, aggregation, and sparsity recognition performance. Regarding estimation accuracy, we compute the Frobenius norm $\| \ma{\Theta} - \ma{\hat{\Theta}} \|_{F}$. For aggregation performance, we report the estimated number of clusters $\hat{K}$ as well as the adjusted Rand index (ARI) \citep{hubert1985} of the obtained clustering of variables. Note that $\hat{K}$ directly follows from the obtained clustering for the optimal values of the tuning parameters from cross-validation.
For sparsity recognition, we report the false positive and false negative rates (FPR and FNR, respectively), where the FPR reports the fraction of true zero elements of the precision matrix that are estimated as nonzero and the FNR gives the fraction of true nonzero elements of the precision matrix that are estimated as zero.

\subsection{Results}
\label{sec:sim-results}
\begin{figure}[!t]
\centering
\includegraphics[width=0.86\textwidth]{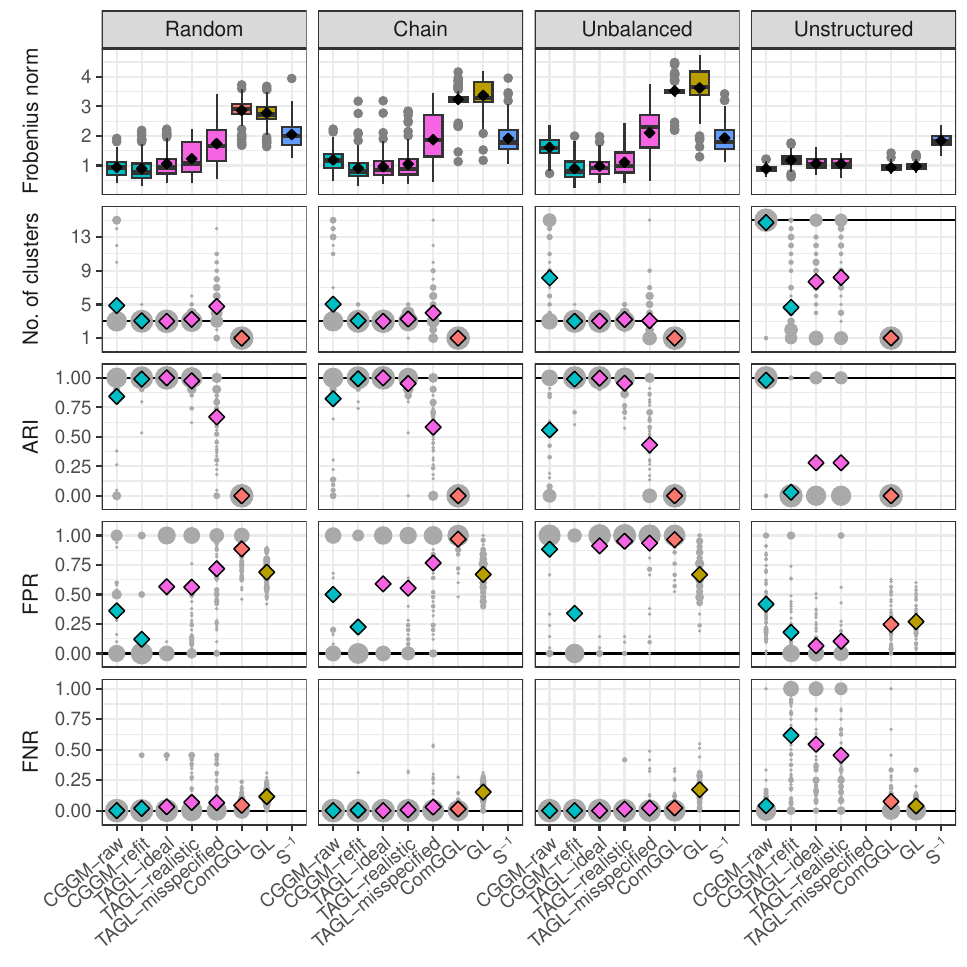} \vspace{-10pt}
\caption{Results for the baseline simulation designs (columns). Top row: Boxplots of the Frobenius norm with black diamonds representing the average. Other rows: Diamonds displaying the average of the estimated number of clusters, ARI, FPR, and FNR. Reference lines are added for the true number of clusters, the ARI value of perfect clustering, and the FPR and FNR of perfect sparsity recognition, respectively. The size of the grey dots represents the frequency of different values across the replications.
Aggregation performance is not applicable and omitted for GL and $\ma{S}^{-1}$, as is sparsity recognition performance for $\ma{S}^{-1}$.
In the unstructured design, a misspecified tree for TAGL does not exist since any tree hierarchy contains the true clustering (each variable being its own cluster).}
\label{fig:WB2022_baseline}
\end{figure}

\begin{figure}[!t]
\centering
\includegraphics[width = 0.86\textwidth]{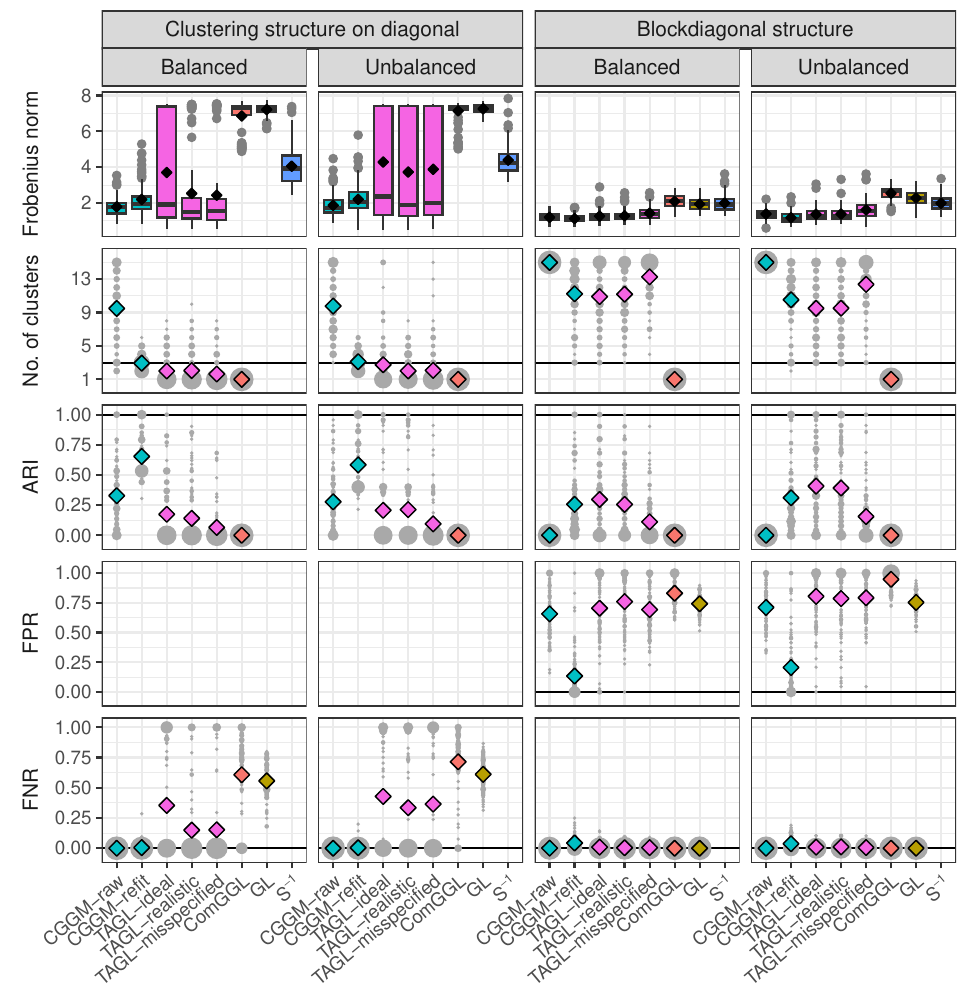}\vspace{-10pt}
\caption{Results for the simulation designs with clustering structure on the diagonal (left) and with a blockdiagonal structure (right). See Figure \ref{fig:WB2022_baseline} for explanatory notes. The FPR is not applicable in the design with clustering structure on the diagonal, since no elements of the true precision matrix are zero.}
\label{fig:diagonal_blockdiagonal}
\end{figure}

In Figures~\ref{fig:WB2022_baseline} and \ref{fig:diagonal_blockdiagonal}, we report the evaluation criteria for different simulation designs.
As CGGM and TAGL outperform ComGGL, the graphical lasso (GL), and the inverse of the sample covariance matrix ($\ma{S}^{-1}$) in almost all cases, we focus on the former two in the following discussion of the results.

{\bf Baseline simulation designs.} In terms of estimation accuracy and aggregation, CGGM-raw and CGGM-refit perform similar to TAGL-ideal, outperform TAGL-realistic in some designs, and outperform and TAGL-misspecified in all designs (see Figure~\ref{fig:WB2022_baseline}).
Regarding sparsity recognition, one variant of CGGM generally outperforms all variants of TAGL, depending on the setting.
The refitting step of CGGM is advantageous in the three block designs but disadvantageous in the unstructured design. For the latter, the refitting step often yields only one or two clusters, likely due to the high degree of sparsity in the unstructured design. This pooling of estimates for the large number of zero elements reduces noise, potentially improving the out-of-sample cross-validation score in Equation~\eqref{eq:cv_score}, despite introducing some bias in the relatively small number of nonzero elements. Conversely, CGGM-raw allows for shrinkage between elements without aggregation, thus improving estimation performance for designs in which many off-diagonal elements share the same value without belonging to the same cluster.
While TAGL-misspecified often selects the correct number of clusters, the ARI reveals that it struggles to accurately classify the variables within those clusters due to the misspecified tree. This pattern persists across most of the simulation designs.
Note that across all simulations, ComGGL detects a single community.
This should, however, not be interpreted as a block-structured precision matrix estimate with a single cluster in the sense of Equation~\eqref{theta-block-form}, as ComGGL allows entries of the precision matrix estimate within the same community to vary.

{\bf Increasing the number of variables or clusters.}
Given that CGGM-refit clearly outperforms \mbox{CGGM-raw} in the baseline chain design, we omit the results for the latter in the variations with increasing number of variables or clusters.
The results remain relatively stable when increasing the number of variables or clusters and are similar to those in the baseline setting (see Figures~\ref{fig:WB2022_variables} and~\ref{fig:WB2022_clusters} in Appendix~\ref{appendix:simulation}), likely due to maintaining a fixed ratio of $n$ to $p$. CGGM-refit performs comparably to TAGL-ideal and TAGL-realistic in terms of estimation accuracy and aggregation, and clearly surpasses TAGL-misspecified.
In terms of sparsity recognition, CGGM-refit and all variants of TAGL have a near-perfect FNR, but CGGM-refit has a lower FPR. This finding remains stable with an increasing number of variables (except for some variation in the FPR of TAGL-realistic and TAGL-misspecified). For an increasing number of clusters, the variants of TAGL exhibit a more prominent improvement in FPR than CGGM-refit, although the latter remains better.

{\bf Approximate block structure.} The results for estimation accuracy are highly similar to those of the baseline designs (see Figure \ref{fig:imperfect} in Appendix~\ref{appendix:simulation}). In terms of aggregation performance, CGGM-raw struggles to find the block structure when it is no longer exact, but the refitting step overcomes this issue. TAGL-realistic also performs worse in terms of aggregation, though the effect is less pronounced than with CGGM-raw. Overall, aggregation performance of CGGM-refit and TAGL-ideal remains very similar to the baseline designs. It is important to note that, technically, the identified block structure is incorrect at the population level. However, the noise reduction in finite samples is still beneficial. Concerning sparsity recognition, the FNR of all variants of CGGM and TAGL stays near-perfect whereas the FPR generally increases. CGGM-refit thereby shows the smallest increase in FPR.

{\bf Clustering structure on diagonal.} CGGM-refit clearly outperforms the alternative methods in terms of aggregation performance (see left two columns in Figure \ref{fig:diagonal_blockdiagonal}). All variants of TAGL tend to aggregate all variables into one cluster, as they exclude the elements on the diagonal from the aggregation penalty. As a result, the variability in the estimation error of TAGL is considerably larger compared to CGGM. Despite the better aggregation performance, the estimation error is slightly higher for CGGM-refit than for CGGM-raw.
A likely explanation of this phenomenon is that CGGM-refit no longer applies shrinkage to the difference between elements from different clusters. In this design, this behavior does not align with the structure of the underlying precision matrix, in which all off-diagonal elements share the same value. Consequently, by shrinking the off-diagonal elements from different clusters towards each other, CGGM-raw achieves superior estimation accuracy even without finding the true cluster structure. In addition, the two variants of CGGM yield a near-perfect and lower FNR than all three variants of TAGL.

{\bf Blockdiagonal structure.}
While CGGM-refit shows comparable estimation accuracy and aggregation performance to TAGL-ideal and TAGL-realistic, it demonstrates a sizable advantage over its alternatives in terms of sparsity recognition due to a much lower FPR
(see right two columns in Figure \ref{fig:diagonal_blockdiagonal}).
As in the designs with an approximate block structure, the identified block structure is technically incorrect at the population level. Nonetheless, identifying groups remains beneficial, considering the limited number of nonzero elements outside the diagonal blocks.
Furthermore, it is noteworthy that CGGM-raw achieves competitive estimation accuracy with minimal aggregation.

{\bf Conclusions from the simulations.} CGGM-refit exhibits excellent performance similar to TAGL-ideal while not requiring auxiliary information. Additionally, the simulations reveal that a misspecified tree that encodes the aggregation is detrimental to the performance of TAGL, positioning CGGM as a particularly strong alternative when accurate information on the aggregation structure is unavailable.
CGGM consistently surpasses or at least keeps up with alternatives across the investigated simulation designs.
In general, the refitting step is beneficial, but may lead to overaggregation and overly sparse estimates in unstructured settings.
Correspondingly, in some scenarios, we find that there is a trade-off between CGGM-raw and CGGM-refit in terms of estimation accuracy, aggregation performance, and sparsity recognition.

{\bf Computation time.} We compare the computation time of CGGM, TAGL, and ComGGL using the same data generating process as in the simulation design with increasing number of variables, which keeps the ratio $n/p$ constant. For each method, we compute the total computation time over a grid of values for the aggregation parameter as used in the simulations, averaged over 10 replications. Other tuning parameters are fixed to values that were frequently found to be optimal (the fifth value in the grid for the sparsity parameter for all methods; $k = 5$ for the weight matrix in CGGM). Figure~\ref{fig:computation_time} in Appendix~\ref{appendix:simulation} displays the resulting computation times. ComGGL scales best as the number of variables increases. However, CGGM both has lower initial computation time and scales better than TAGL, thereby offsetting the additional computational burden for tuning additional hyperparameters.

\section{Estimation of a Clustered Covariance Matrix}
\label{sec:estimate_cov}

While we focused so far on estimating a clustered precision matrix to simplify the partial dependency structure among the variables, it can also be of interest to obtain a clustered covariance matrix. For instance, a block structure in the covariance matrix implies that the variables of a given block have equal loadings in latent factor models. In this context, a clustered covariance matrix may reduce uncertainty in estimating the latent factors.

\begin{figure}[!t]
\centering
\includegraphics[width = 0.8\textwidth]{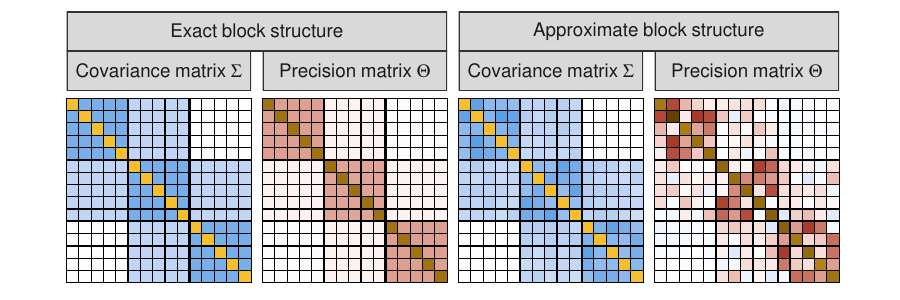} \vspace{-12pt}
\caption{Illustration of how an exact block structure is retained between the covariance matrix $\ma{\Sigma}$ and the precision matrix $\ma{\Theta}$ (left), whereas an approximate block structure in $\ma{\Sigma}$ corresponds to a much more noisy structure in $\ma{\Theta}$ (right).}
\label{fig:covariance_precision}
\end{figure}

Existing paradigms for regularized estimation either induce an appropriate simplicity structure on the precision matrix or the covariance matrix, and the induced simplicity structure is, in general, not maintained when taking the inverse of the object of interest (e.g., \citealp{yao2019clustered}, \citealp{pircalabelu2020community}, and \citealp{wilms2021tag_lasso}).
The proposed CGGM estimator connects both paradigms since it retains the found block structure when taking the inverse of the obtained estimate (Section \ref{subsec:GGM}).
If model~\eqref{theta-block-form} holds for the precision matrix~$\ma{\Theta}$, it may therefore be reasonable to obtain a clustered estimate of the covariance matrix $\ma{\Sigma}$ by taking $\ma{\hat{\Sigma}} = \ma{\hat{\Theta}}^{-1}$, where $\ma{\hat{\Theta}}$ denotes the CGGM estimate of the precision matrix.
While the induced shared block structure in the precision and covariance matrix is a distinct feature of our proposal,
adherence to an exact block structure as in model~\eqref{theta-block-form} may, however, be a strong assumption in practice. If this model only holds approximately with somewhat similar values within the blocks of $\ma{\Theta}$, we have seen in the simulations that estimating a clustered precision matrix with CGGM is nevertheless beneficial. But if an approximate block structure is present in the covariance matrix $\ma{\Sigma}$, the structure in $\ma{\Theta}$ is much more noisy with larger variability between the elements in the corresponding blocks, see  Figure~\ref{fig:covariance_precision}. In such settings, first estimating a clustered precision matrix with CGGM and then taking the inverse may not succeed in finding the correct structure.

To estimate a clustered covariance matrix, one could attempt to adapt the clusterpath algorithm to be based on the objective function
\begin{equation*}
    -\log |\ma{\Sigma}^{-1}| + \tr \ma{S} \ma{\Sigma}^{-1} + \lambda \mathcal{P}(\ma{\Sigma}).
\end{equation*}
An algorithm optimizing this objective function must deal with an additional layer of complexity, as evaluation of $\ma{\Sigma}$ is required in the penalty part and evaluation of $\ma{\Sigma}^{-1}$ in the likelihood part of the objective function.

As an alternative, consider a random vector $\ve{X}$ that follows a normal distribution with covariance matrix~$\ma{\Sigma}$. Then there exists an affine transformation matrix $\ma{A} = \ma{A}(\ma{\Sigma})$ so that $\ve{Y} = \ma{A}^\top \ve{X}$ follows a normal distribution with covariance matrix $\ma{\Theta} = \ma{\Sigma}^{-1}$. That is, $\ma{\Sigma}$ is the precision matrix for the random vector $\ve{Y}$, which implies that we can apply CGGM to minimize the objective function $L(\ma{\Sigma})$ from~\eqref{eq:cgm_opt_problem} with the following minor modification. As we do not observe realizations of $\ve{Y}$, we cannot compute the sample covariance matrix of those realizations. Instead, we take the inverse of the sample covariance matrix $\ma{S}^{-1}$ of the observed realizations of $\ve{X}$ as input for the CGGM algorithm. The corresponding optimization problem remains convex, but practitioners can benefit---in terms of estimation accuracy and interpretability---from tweaking the optimization problem towards directly solving for the covariance matrix instead of the precision matrix.

\begin{figure}[!t]
\centering
\includegraphics[width = 0.99\textwidth]{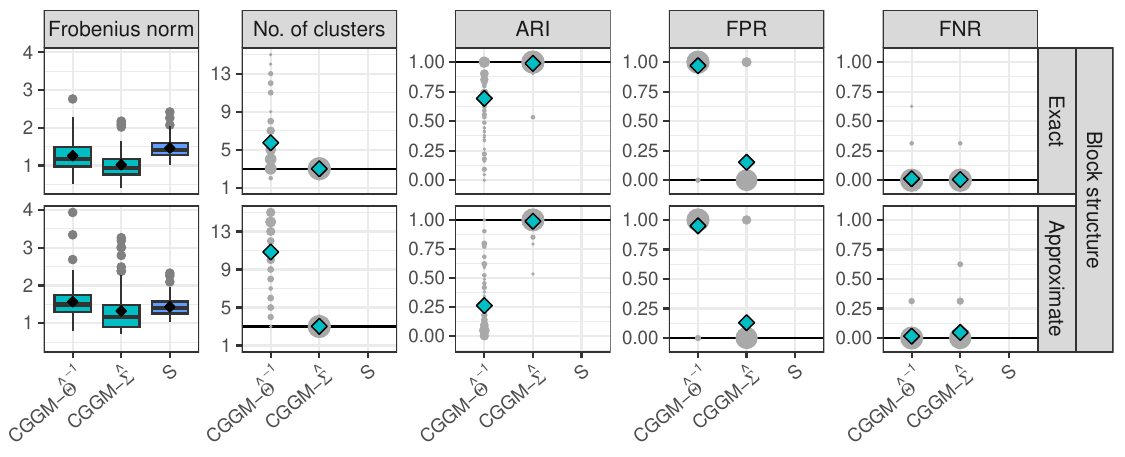} \vspace{-10pt}
\caption{Simulation results for the estimation of a clustered covariance matrix, with different simulation designs in separate rows and different evaluation metrics in separate columns. Cf.\ Figure~\ref{fig:WB2022_baseline} for explanatory notes.}
\label{fig:Sigma}
\end{figure}

To demonstrate this numerically, we conduct simulations using the baseline chain design---which employs an exact block structure---and the approximate block structure chain design (see Section~\ref{sec:sim-design}), but with the block structure in the covariance matrix $\ma{\Sigma}$ rather than the precision matrix $\ma{\Theta}$. We apply CGGM for obtaining a clustered precision matrix followed by taking the inverse (denoted by CGGM-$\ma{\hat{\Theta}}^{-1}$) and the modification for obtaining a clustered covariance matrix (denoted by CGGM-$\ma{\hat{\Sigma}}$), as well as the sample covariance matrix $\ma{S}$. Both variants of CGGM include the parameter re-estimation step from Section~\ref{subsubsec:refit}.
As the tuning parameter $\phi$ may have a different effect on clustering the covariance and precision matrix, respectively, we now also include candidate values $\phi \in \{1, 2, 3\}$ in the cross-validation scheme.
The results are shown in Figure~\ref{fig:Sigma}. Clearly, CGGM-$\ma{\hat{\Sigma}}$ succeeds in finding the relevant aggregation and sparsity structure in most replications whereas CGGM-$\ma{\hat{\Theta}}^{-1}$ struggles to do so, even for the setting with an exact block structure. While both variants of CGGM improve upon the sample covariance matrix $\ma{S}$ in terms of the estimation error, CGGM-$\ma{\hat{\Sigma}}$ exhibits the lowest error. Although a more thorough evaluation is beyond the scope of this paper, our findings indicate that applying CGGM-$\ma{\hat{\Sigma}}$ may be preferable over CGGM-$\ma{\hat{\Theta}}^{-1}$ when a clustered covariance matrix is of primary interest.
We therefore recommend users to first decide on the object of interest, a block-structured covariance or precision matrix, and then use CGGM accordingly.

\section{Applications}
\label{sec:applications}
We demonstrate CGGM's practical usefulness and versatility on 3 applications: a finance application using stock data  (Section~\ref{subsec:SP100}), an application with country-level well-being indicators (Section~\ref{subsec:wellbeing}), and a survey-based behavioral science application  (Section~\ref{subsec:survey}).

\subsection{S\&P 100 Stocks}
\label{subsec:SP100}
We consider a financial data set containing daily realized ranges---a volatility measure---of $p=101$ stocks of the companies constituting the S\&P 100 on September 18\textsuperscript{th}, 2023.\footnote{The S\&P 100 contains 100 companies that may have one or more classes of stocks listed. During the time frame of our analysis, Alphabet was the sole company with more than one class, distinguished by the symbols GOOG and GOOGL, therefore resulting in $p=101$ instead of $p=100$ variables in our analysis.}
We study the conditional dependency structure of the stocks' realized ranges over the period January 3\textsuperscript{rd}, 2023, until December 29\textsuperscript{th}, 2023 (\mbox{$n=250$}), thereby comparing the performance of CGGM to that of TAGL applied to the precision matrix.
While the former learns how to cluster the variables in an unsupervised and data-driven manner, the latter requires side-information in the form of a tree that encodes the similarity between the stocks to do so.
To this end, we use the Global Industry Classification Standard (GICS). The tree then consists of the $p=101$ stocks (leaves), 11 sectors as middle layer where each stock/company belongs to one industry sector, and one root node that aggregates all sectors.

Since the data are temporally dependent, we first---as a preprocessing step---fit the popular heterogeneous autoregressive (HAR) model of \cite{corsi2009simple} to the individual daily realized ranges to capture time dependencies, and then apply CGGM and TAGL to the standardized residuals to learn the conditional dependency structure among the stocks.
See Appendix~\ref{appendix:SP100} for details on data preprocessing and tuning parameter selection, including the selection of candidate values for $\phi$ based on the resulting weight distributions.

\begin{figure}[!t]
\centering
\includegraphics[width=0.99\textwidth]{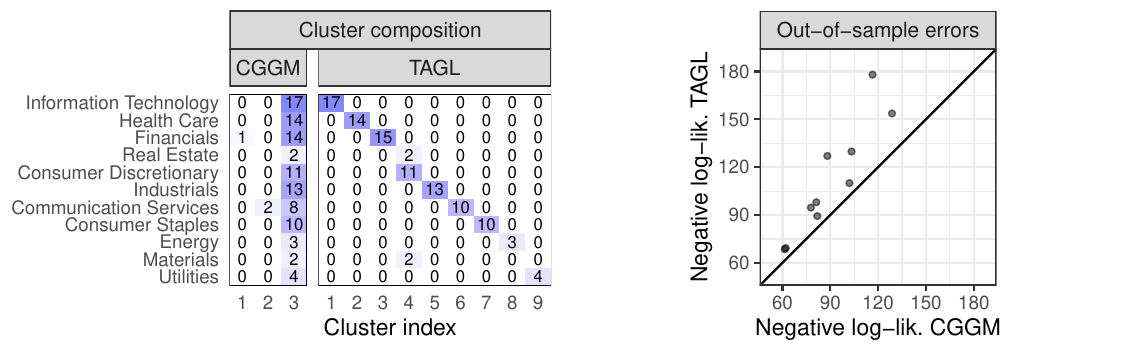} \vspace{-10pt}
\caption{Finance application. Sector distribution over the clusters of CGGM and TAGL (left). Out-of-sample errors across the 10 replications (dots) for CGGM and TAGL (right; note that two of the dots are overplotted).}
\label{fig:finance}
\end{figure}

The cluster solution returned by both procedures differs considerably. CGGM groups the stocks into $K=3$ clusters: the first  contains BAC (issued by Bank of America Corp.), the second  contains GOOG and GOOGL (both issued by Alphabet), and the last cluster contains the remaining stocks.
CGGM thus puts the Bank of America and the technology company Alphabet in the spotlight as opposed to the rest of the market.
TAGL, in contrast, results in $K=9$ clusters where the industry sectors are assigned to the different clusters on an almost one-to-one basis.
In Figure~\ref{fig:finance} (left panel), we present the sector distribution  over the clusters for  CGGM and TAGL. The unbalanced cluster solution of the former versus the more balanced one of the latter directly stands out.
The role of  BAC, GOOG, and GOOGL in our analysis is in line with the central, market-wide role of banks  and technology companies  as their influence on prices affects stocks across the whole market.

Given the considerable difference in clusters returned by both methods, a natural question is how both perform in capturing the conditional dependency structure.
To this end, we conduct an out-of-sample exercise.
An additional outer loop for 10-fold cross-validation is used to compute out-of-sample errors on each of the $G=10$ test samples according to the likelihood-based score \eqref{eq:cv_score}.
These errors on the test data are visualized in Figure~\ref{fig:finance} (right panel) with CGGM on the horizontal  and TAGL on the vertical axis.
For each test sample (dots), CGGM has a lower error thereby indicating that the more unbalanced clusters form a better description of the conditional dependency structure in this context.

\subsection{OECD Well-Being Indicators}
\label{subsec:wellbeing}
The second data set we analyze is one on OECD well-being indicators, collected in 2018 by \cite{cavicchia2022gaussian} (see Appendix~\ref{appendix:OECD}).
The data contain $p=11$ variables related to well-being: education, jobs, income, safety, health, environment, civic engagement, \mbox{accessibility} to services, housing, community, and life satisfaction, on which two groups of countries with  sample sizes $n_1 = 21$ and $n_2 = 14$ are given a score ranging from 0 to 10.

\begin{figure}[!t]
\centering
\includegraphics[width= 0.99\textwidth]{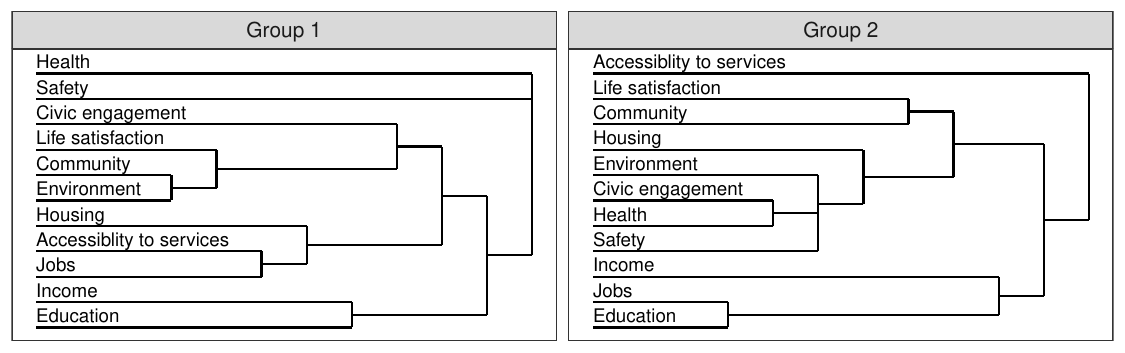} \vspace{-10pt}
\caption{Well-being application. Clusterpath dendrograms obtained by CGGM on the two groups of countries.}
\label{fig:oecd_dendrograms}
\end{figure}

We highlight the capabilities of CGGM in estimating  a cluster hierarchy on the standardized well-being data.
Since our main goal is the retrieve the cluster hierarchy, we set $\lambda_s=0$ and report the whole clusterpath solution  obtained by applying CGGM to the precision matrix with different values of the tuning parameter $\lambda_c$ for the two country groups.
Due to the small group sample sizes, we do not use cross-validation to select the tuning parameters $\phi$ and $k$  but use fixed values based on existing conventions, namely $\phi=0.5$ and $k=3$ (e.g., \citealp{chi2015ama, wang2018sparse}).
Since our primary interest is in the clusterpaths, the refitting step described in Section~\ref{subsubsec:refit} is not required as it does not affect the obtained paths.
Figure~\ref{fig:oecd_dendrograms} visualizes the dendrograms as obtained from CGGM's complete clusterpath solution ranging from $p=11$ clusters to one. If selection of a single solution along the clusterpath is requested, we advice practitioners to resort to domain expertise instead of data-driven methods when the sample size is that limited.

The analysis of the well-being indicators across the two groups of countries reveals distinct clustering patterns that highlight differences in their socioeconomic conditions.
In the first group (e.g., Australia, France, Switzerland), the aggregation of civic engagement, life satisfaction, community, environment, housing, accessibility to services, and jobs appears to reflect the social and environmental dimensions of well-being. Conversely, in the second group (e.g., Chile, Hungary, Mexico), safety and health are combined with the variables representing social and environmental dimensions, possibly suggesting a greater reliance on the community for these services. This contrasts with more developed countries where such services may be perceived as more self-evident. Additionally, in the second group, income, jobs, and education are grouped together, reflecting their importance to socioeconomic progress. Finally, note that we use CGGM with the precision matrix as object of interest; its block structure directly transfers to the covariance matrix thereby making our results comparable to those of \cite{cavicchia2022gaussian} who estimate the clustering structure on the covariance matrix.

\subsection{Humor Styles Questionnaire} \label{subsec:survey}
In behavioral science surveys, multiple rating-scale items are typically used to measure each latent construct of interest. For further analysis, mean scores across the respective items are commonly computed as measurements of each latent construct, which implies the assumption that the covariance matrix of the questionnaire items follows a block structure.
The humor styles questionnaire (HSQ) \citep{martin2003HSQ} measures four latent constructs corresponding to different styles of humor: \emph{affiliative}, \emph{self-enhancing}, \emph{aggressive}, and \emph{self-defeating}, with each construct being measured by eight items ($p = 32$).
If we can retrieve these four clusters of items in the covariance matrix, the assumption behind computing mean scores can be considered reasonable.
Details on the HSQ data and tuning parameter selection in CGGM for clustering the covariance matrix are given in Appendix~\ref{appendix:HSQ}.

CGGM returns four clusters of items that largely overlap with the grouping of \citet{martin2003HSQ}, as shown in Table~\ref{tab:HSQ-clusters}. Looking at the wording of the questionnaire items, the clustering found by CGGM is intuitive. All items of the \emph{affiliative} group, which CGGM places in Cluster 1, refer to humor in social settings. In addition, the two items of the \emph{self-enhancing} group that complete Cluster 1---items~6 and~30---address being alone as opposed to being with people. On the other hand, the remaining six items of the \emph{self-enhancing} group, which CGGM puts in Cluster 2, all refer to using humor for dealing with feeling depressed, upset, unhappy, sad, or having problems. Item 28 from the \emph{self-defeating} group, which completes Cluster 2, also addresses having problems or feeling unhappy. In contrast, the other seven items of this group, which form Cluster 4 in the CGGM solution, refer to laughing or letting others laugh at oneself. Finally, all items from the \emph{aggressive} group address teasing and offensive or inappropriate humor, and are placed in Cluster 3.

\begin{table}[!t]
\linespread{1}\selectfont
\setlength{\tabcolsep}{8pt}
\centering
\begin{tabular}{l p{2.55cm} p{2.55cm} p{2.55cm} p{2.55cm}}
\noalign{\smallskip}\toprule\noalign{\smallskip}
 & Cluster 1 & Cluster 2 & Cluster 3 & Cluster 4 \\
\noalign{\smallskip}\midrule\noalign{\smallskip}
Affiliative & \{1, 5, 9, 13, 17, 21, 25, 29\} & & & \\
\noalign{\smallskip}
Self-enhancing & \{6, 30\} & \{2, 10, 14, 18, 22, 26\} & & \\
\noalign{\smallskip}
Aggressive & & & \{3, 7, 11, 15, 19, 23, 27, 31\} & \\
\noalign{\smallskip}
Self-defeating & & \{28\} & & \{4, 8, 12, 16, 20, 24, 32\} \\
\noalign{\smallskip}\bottomrule
\end{tabular}
\caption{Humor styles questionnaire application. Allocation of the items for measuring the four humor styles (in rows) to the four clusters found by CGGM (in columns).}
\label{tab:HSQ-clusters}
\end{table}

Regarding sparsity, \citet[Table~2]{martin2003HSQ} report significant positive correlations between most of the four humor styles, at least for male participants. CGGM on the given data, on the other hand, finds nonzero covariance only between Clusters~1 and~2 as well as Clusters~2 and~4. That is, through regularization, the structure found by CGGM suggests latent variables that are more distinct than the factors found by \citet{martin2003HSQ}, which is desirable when characterizing different humor styles.

Although the above interpretation of the obtained clusters is subjective, our findings suggest that the survey design of the HSQ may benefit from further finetuning. Moreover, if subsequent analysis is based on mean scores, a slightly different allocation of the items to the latent constructs---the humor styles---may be more in line with the implicit assumption of a block structure in the covariance matrix.

\newpage
\section{Conclusion}
\label{sec:conclusion}
We introduce a novel method to estimate Gaussian graphical models (GGMs) subject to node-clustering in addition to edge-sparsity. Our method, which we call CGGM,
uses a clusterpath penalty to produce a hierarchical clustering of variables (nodes in the graph) without relying on pre-existing notions of the cluster composition.
Our software package \pkg{clusterGGM} \citep{clusterGGM} for the statistical computing environment \proglang{R} \citep{R} implements the proposed method and is available from CRAN (the Comprehensive \proglang{R} Archive Network) at \url{https://CRAN.R-project.org/package=clusterGGM}.

In a comprehensive simulation study covering a wide range of graph structures, we compare CGGM to similar benchmarks such as TAGL, ComGGL, and the graphical lasso. CGGM oftentimes surpasses the benchmarks in terms of estimation accuracy as well as clustering and sparsity recognition performance, though the latter require a refitting step particularly in noisy simulation designs.
Through a diverse set of applications, we also demonstrate the versatility of CGGM in
(i) learning cluster hierarchies in a fully data-driven way as compared to other benchmarks such as TAGL that require additional side-information for this task,
(ii) delivering accurate cluster hierarchies that are transferable between the precision and covariance matrix, and
(iii) directly estimating clustered covariance matrices if these are the primary object of interest.

Concerning future work, CGGM may be extended to partial correlation matrices instead of precision matrices. In contrast to the graphical lasso (GL), \cite{carter2024partial} apply the sparsity penalty to the elements of the partial correlation matrix and call their method the partial correlation GL. As a result, the sparsity structure estimated by this method is invariant to scalar multiplication of the data. CGGM can also be modified to accommodate clustering based on the partial correlation matrix, but this comes at the cost of the convexity of the objective function as with the partial correlation GL.

\section*{Computational Details}
All computations were performed with \proglang{R}
\citep{R}. Replication files are available from \url{https://github.com/aalfons/CGGM-replication}.

\section*{Acknowledgments}
Andreas Alfons is supported by a grant from the Dutch Research Council (NWO), research program Vidi, grant number VI.Vidi.195.141.
Ines Wilms is supported by the same funding agency under grant number VI.Vidi.211.032.

\bibliographystyle{apalike2}
\bibliography{references}

\newpage
\appendix

\section{Derivations for the Clusterpath Estimator} \label{app:clusterpathderivatation}

\subsection{Re-expressing the Objective Function $L(\ma{\Theta})$}\label{app:Ltheta_reexpressed}
We detail how the objective function in terms of $\ma{\Theta}$, namely
\begin{equation}
L(\ma{\Theta}) = -\log | \ma{\Theta} | + \tr \ma{S\Theta} + \mathcal{P}(\ma{\Theta}), \label{app:objective_function}
\end{equation}
can be re-expressed in terms of ($\ma{A}, \ma{R}$) by using the $G$-block format
$\ma{\Theta} = \bf{U} \bf{R} \bf{U}^\top + {\bf A}$ for a certain number of clusters $K$,
where $\bf{U}$ is the membership matrix with $u_{jk} = 1$ if variable $j$ belongs to cluster $k$ and zero otherwise, $\ma{R}=(r_{k\ell})_{1 \leq k, \ell \leq K}$ is a symmetric matrix, and ${\bf A} = \text{diag}(a_{11} \cdot \ma{I}, \ldots, a_{KK}\cdot\ma{I})$ is a diagonal matrix.
The main purpose of this re-expression
is to obtain a more compact expression that permits computationally efficient updates for solving the optimization problem
\begin{align}
    \widehat{\ma{\Theta}} &= \argmin_{\ma{\Theta}} \ L(\ma{\Theta}) \qquad \text{s.t.} \  \ma{\Theta}=\ma{\Theta}^\top, \ma{\Theta} \succ 0. \nonumber
\end{align}
To this end, we begin by defining the key variables and then work our way through the different terms of the objective function \eqref{app:objective_function}.

For any number of clusters $1 \leq K \leq p$, the precision matrix displays the block-structure
\begin{equation}
\label{app:eq:theta_decomposed}
\ma{\Theta} =
 \begin{bmatrix}
     a_{11}\ma{I} + r_{11} \ma{11}^\top & r_{12} \ma{11}^\top & \cdots &r_{1K} \ma{11}^\top \\
      r_{21} \ma{11}^\top & a_{22}\ma{I} + r_{22} \ma{11}^\top & \cdots & r_{2K} \ma{11}^\top \\
      \vdots & \vdots & \ddots & \vdots \\
     r_{K1} \ma{11}^\top & r_{K2} \ma{11}^\top & \cdots & a_{KK}\ma{I} + r_{KK} \ma{11}^\top \\
  \end{bmatrix}.
\end{equation}
Without loss of generality, we assume that the variables have been reordered such that those in cluster 1 form the first block, those in cluster 2 the second block, and so on.
Let $\ma{P} = \ma{U}^\top\ma{U}$ be the diagonal matrix with the number of variables $p_k$ per cluster as diagonal elements,
$\ma{P}^{1/2}$ then simply contains their square roots,
and $\ma{p} = \ma{1}^\top \ma{U}$ is the $K \times 1$ vector with elements $p_k$. It can be verified that $\ma{X}\ma{U}\ma{P}^{-1}$ computes the cluster averages of the original variables in $\ma{X}$.

\paragraph{Log determinant term.} First, we focus on the first term in objective function \eqref{app:objective_function}, the log determinant of $\ma{\Theta}$, to obtain a convenient expression in $\ma{A}$ and $\ma{R}$. We show that $\ma{\Theta}$ can be split in a linear space that depends on $\ma{A}$ and $\ma{R}$ and an orthogonal linear space that is only dependent on the diagonal blocks of $\ma{\Theta}$ and thus on $\ma{A}$ and the diagonal elements of $\ma{R}$. To do so, we first consider what happens if we take cluster averages, that is,
\begin{align*}
    \ma{P}^{-1}\ma{U}^\top \bm{\Theta} \ma{U}\ma{P}^{-1}
    &= \ma{P}^{-1}\ma{U}^\top \left(  \ma{URU}^\top +
    \begin{bmatrix}
        a_{11}\ma{I}  & \ma{0} & \cdots & \ma{0} \\
        \ma{0} & a_{22}\ma{I} & \cdots & \ma{0} \\
        \vdots & \vdots & \ddots & \vdots \\
        \ma{0} & \ma{0} & \cdots & a_{KK}\ma{I} \\
    \end{bmatrix}
    \right)
    \ma{U}\ma{P}^{-1} \\
    &= \ma{R} + \ma{P}^{-1} \ma{A} \\
    & = \ma{R}^*.
\end{align*}
Then,
\begin{equation*}
    \ma{UR}^*\ma{U}^\top =
    \begin{bmatrix}
        r_{11}^* \ma{11}^\top & r_{12} \ma{11}^\top & \cdots & r_{1K} \ma{11}^\top \\
        r_{21} \ma{11}^\top & r_{22}^* \ma{11}^\top & \cdots & r_{2K} \ma{11}^\top \\
        \vdots & \vdots & \ddots & \vdots \\
        r_{K1} \ma{11}^\top & r_{K2} \ma{11}^\top & \cdots & r_{KK}^* \ma{11}^\top \\
    \end{bmatrix},
\end{equation*}
where $r_{kk}^* = a_{kk}/p_k + r_{kk}$ so that $\bm{\Theta} - \ma{UR}^*\ma{U}^\top$ has off-diagonal blocks equal to zero and diagonal blocks equal to $a_{kk}(\ma{I} - p_k^{-1} \ma{11}^\top) = a_{kk} \ma{J}_k$ with $\ma{J}_k$ the $p_k \times p_k$ centering matrix. This yields
\begin{equation}
    \ma{\Theta} = \ma{UR}^*\ma{U}^\top +
    \begin{bmatrix}
        a_{11} \ma{J}_1 & \ma{0}  & \cdots & \ma{0} \\
        \ma{0}  & a_{22} \ma{J}_2 & \cdots & \ma{0} \\
        \vdots & \vdots & \ddots & \vdots \\
        \ma{0}  & \ma{0}  & \cdots & a_{KK} \ma{J}_K \\
    \end{bmatrix}. \label{app:theta-decomposed}
\end{equation}
This presents an orthogonal decomposition where the precision matrix is split in two parts, the first part is a linear space that depends on $\ma{A}$ and all elements of $\ma{R}$, the second part is an orthogonal linear space that only depends on the diagonal blocks of the precision matrix, namely $\ma{A}$.
It can be verified that this decomposition is indeed orthogonal because post-multiplying the final term with $\ma{U}$ yields zero due to the centering matrices.

Next, it is well known that the determinant is equal to the product of the eigenvalues of the matrix. Therefore, we require the eigenvalues of both parts of decomposition \eqref{app:theta-decomposed}. Let the eigendecomposition of $\ma{\Theta}$ be given by
\begin{align*}
    \ma{\Theta} &= \ma{Q}\ma{\Gamma}\ma{Q}^\top \\
    &= \ma{Q}_1\ma{\Gamma}_1\ma{Q}_1^\top + \ma{Q}_2\ma{\Gamma}_2\ma{Q}_2^\top \\
    &= \ma{UR}^*\ma{U}^\top +
    \begin{bmatrix}
        a_{11} \ma{J}_1 & \ma{0} & \cdots & \ma{0}  \\
        \ma{0}    & a_{22} \ma{J}_2 & \cdots &\ma{0}  \\
        \vdots & \vdots & \ddots & \vdots \\
        \ma{0}  & \ma{0}  & \cdots &  a_{KK} \ma{J}_K \\
    \end{bmatrix}.
\end{align*}
We now need to find $\ma{\Gamma}_1$ in the eigendecomposition $\ma{UR}^*\ma{U}^\top = \ma{Q}_1\ma{\Gamma}_1\ma{Q}_1^\top$. Pre- and post-multiplying with an orthonormal matrix does not change the eigenvalues. A convenient orthonormal matrix for pre-multiplication is therefore $\ma{P}^{-1/2}\ma{U}$ giving
\begin{align*}
    \ma{UR}^*\ma{U}^\top &= \ma{Q}_1\ma{\Gamma}_1\ma{Q}_1^\top \\
    \ma{P}^{-1/2}\ma{U}^\top\ma{UR}^*\ma{U}^\top\ma{U}\ma{P}^{-1/2} &= \ma{P}^{-1/2}\ma{U}^\top\ma{Q}_1\ma{\Gamma}_1\ma{Q}_1^\top\ma{U}\ma{P}^{-1/2} \\
    \ma{P}^{1/2}\ma{R}^*\ma{P}^{1/2} &= \ma{Q}_1^*\ma{\Gamma}_1{\ma{Q}_1^*}^\top \\
    \begin{bmatrix}
        p_1 r_{11}^*  & (p_1 p_2)^{1/2}r_{12} & \cdots & (p_1 p_K)^{1/2}r_{1K} \\
        (p_1 p_2)^{1/2}r_{21}  & p_2 r_{22}^* & \cdots & (p_2 p_K)^{1/2}r_{2K} \\
        \vdots & \vdots & \ddots & \vdots \\
        (p_1 p_K)^{1/2}r_{K1}  & (p_2 p_K)^{1/2}r_{K2} & \cdots &  p_K r_{KK}^*  \\
    \end{bmatrix}&= \ma{Q}_1^*\ma{\Gamma}_1{\ma{Q}_1^*}^\top,
\end{align*}
which shows that in the case of $K$ clusters, the eigenvalues corresponding to the space spanned by $\ma{U}\ma{R}^*\ma{U}^\top$ in decomposition \eqref{app:theta-decomposed} can be obtained by considering the eigenvalues of a $K \times K$ matrix.

The second part of the space in decomposition \eqref{app:theta-decomposed} is spanned by the diagonal centering matrices \mbox{$a_{kk} \ma{J}_k$}. The centering matrix is a projector matrix that has $p_k - 1$ eigenvalues of 1, so that $a_{kk} \ma{J}_k$ has $p_k - 1$ eigenvalues of $a_{kk}$.

Combining both parts, we can express
\begin{equation*}
 -\log|\bm{\Theta}| = -\log|\ma{P}^{1/2}\ma{R}^*\ma{P}^{1/2}|
  - \sum_{k=1}^K (p_k - 1)\log a_{kk}.
\end{equation*}
This decomposition shows that the eigenvalues of $\ma{\Theta}$ consist of the eigenvalues of $\ma{P}^{1/2} \ma{R}^* \ma{P}^{1/2}$, together with $a_{kk}$, which appears $p_k-1$ times for each $1\leq k\leq K$. Therefore, if $\ma{R}^*$ is positive definite and $a_{kk}>0$, it follows that $\ma{\Theta}$ is also positive definite.

\paragraph{Trace term.} Second, we address the second term in objective function \eqref{app:objective_function}, the trace of the matrix product $\ma{S\Theta}$. It is straightforward to express this term in terms of $\ma{A}$ and $\ma{R}$, namely
\begin{equation*}
    \tr \ma{S\Theta} = \tr \ma{SURU}^\top + \sum_{\ell=1}^K a_{\ell\ell} \tr \ma{S}_\ell,
\end{equation*}
where $\ma{S}_\ell$ is the sample covariance matrix computed from the $p_\ell$ variables in cluster $\ell$.

\paragraph{Aggregation penalty.} Third, the aggregation penalty term in objective function \eqref{app:objective_function} can be written as
\begin{equation*}
    \sum_{j=2}^p\sum_{j'=1}^{j-1} w_{jj'} d_{jj'}(\bm{\Theta})
    = \sum_{k=1}^K \sum_{\ell=1}^{k-1} \sum_{j \in \mathcal{C}_k} \sum_{j' \in \mathcal{C}_\ell} w_{jj'}  d_{jj'}(\ma{\Theta}),
\end{equation*}
where $\mathcal{C}_k$ denotes the set of variables belonging to cluster $k$.
Because $d_{jj'}(\bm{\Theta})$ evaluates to the same value for all $j \in \mathcal{C}_k$ and $j' \in \mathcal{C}_\ell$, the penalty can be further reduced into
\begin{equation*}
    \sum_{k=1}^K \sum_{\ell=1}^{k-1} \left( \sum_{j \in \mathcal{C}_k} \sum_{j' \in \mathcal{C}_\ell} w_{jj'} \right) d_{\mathcal{C}_k \mathcal{C}_\ell}(\ma{A}, \ma{R}) = \sum_{k=1}^K \sum_{\ell=1}^{k-1} \ma{u}_k^\top \ma{W} \ma{u}_\ell d_{\mathcal{C}_k \mathcal{C}_\ell}(\ma{A},\ma{R}),
\end{equation*}
where $\ma{W}$ is the symmetric $p \times p$ weight matrix containing the individual weights $w_{jj'}$, $\ma{u}_k$ denotes the $k$\textsuperscript{th} column of the membership matrix $\ma{U}$, and
\begin{align*}
    d_{\mathcal{C}_k \mathcal{C}_\ell}^2(\ma{A},\ma{R}) ={}& (a_{kk} + r_{kk} - a_{\ell\ell} - r_{\ell\ell})^2 + (p_k - 1)(r_{kk} - r_{k\ell})^2 + (p_\ell - 1)(r_{\ell\ell} - r_{k\ell})^2 \\
    &+ \sum_{\substack{m=1\\  m \notin \{k, \ell\}}}^K p_m(r_{km} - r_{\ell m})^2.
\end{align*}
Note that the subscripts $\mathcal{C}_k \mathcal{C}_\ell$ in $d_{\mathcal{C}_k \mathcal{C}_\ell}(\ma{A},\ma{R})$ always refer to the cluster formulation of the Euclidean distance.

\paragraph{Sparsity penalty.}
Fourth, consider the sparsity penalty in objective function \eqref{app:objective_function}. By setting $z_{jj}=0$ to avoid penalization of the diagonal elements of $\ma{\Theta}$, the sparsity penalty  can be written as
\begin{equation*}
    \sum_{\substack{j,j' \\ j \neq j'}} z_{jj'}  |\theta_{jj'}| = \sum_{k=1}^K \sum_{\ell=1}^K \sum_{j \in \mathcal{C}_k} \sum_{j' \in \mathcal{C}_\ell} z_{jj'} |\theta_{jj'}|.
\end{equation*}
Using the block structure of $\ma{\Theta}$, this expression can be reformulated in terms of $\ma{R}$ via
\begin{equation*}
    \sum_{k=1}^K \sum_{\ell=1}^K \left( \sum_{j \in \mathcal{C}_k} \sum_{j' \in \mathcal{C}_\ell} z_{jj'} \right) |r_{k\ell}| = \sum_{k=1}^K \sum_{\ell=1}^K \ma{u}_k \ma{Z} \ma{u}_{\ell}^\top  |r_{k\ell}|,
\end{equation*}
where $\ma{Z}$ is the symmetric $p \times p$ weight matrix containing the sparsity weights $z_{jj'}$.
We take the sparsity weights $z_{jj'}$ to be the absolute value of the corresponding entry in  $\ma{S}^{-1}$ or $(\ma{S} + \ma{I})^{-1}$ when the inverse of the sample covariance matrix is not available.

Finally, combining all decomposed terms we obtain the expression
for the objective function $L(\ma{\Theta})$ in terms of $\ma{A}$ and $\ma{R}$, namely
\begin{align}
    L(\ma{\Theta}) ={}&  L(\ma{A}, \ma{R}) \nonumber\\
    ={}&  -\log|\ma{P}^{1/2}\ma{R}^*\ma{P}^{1/2}|
    - \sum_{k=1}^K (p_k - 1)\log a_{kk} + \tr \ma{SURU}^\top \nonumber\\
    &+ \sum_{\ell=1}^K a_{\ell\ell} \tr \ma{S}_\ell + \lambda_c \sum_{k=1}^K \sum_{\ell=1}^{k-1} \ma{u}_k^\top \ma{W} \ma{u}_\ell d_{\mathcal{C}_k \mathcal{C}_\ell}(\ma{A}, \ma{R}) + \lambda_s \sum_{k=1}^K \sum_{\ell=1}^K \ma{u}_k \ma{Z} \ma{u}_{\ell}^\top  |r_{k\ell}|. \label{eq:loss-decomposed}
\end{align}

\subsection{Re-expressing the Objective Function $L(\ma{\Theta}_k)$}
\label{app:subsec:parameterization}
In Section~\ref{app:Ltheta_reexpressed}, efficient expressions $L(\ma{A}, \ma{R})$ for $L(\ma{\Theta})$ have been obtained that make use of the block structure of $\ma{\Theta}$. What remains to be done is to separate $L(\ma{A}, \ma{R})$ into those parts that depend on cluster $k$ and those that do not. This separation is needed to  cycle through the $K$ blocks, to---in turn---update the objective function $L(\ma{\Theta}_k)$ for block/cluster~$k$. In the following, we make the assumption, without loss of generality, that the variables are arranged to consistently position cluster $k$ as the last cluster.

Consider the precision matrix split according to the elements that belong to cluster $k$ ($\ma{\Theta}_k$) and those that do not ($\ma{\Theta}_{-k}$)
\begin{align*}
    \ma{\Theta}
    &=  \ma{\Theta}_{-k} + \ma{\Theta}_k
         = \left[\begin{array}{@{}c|c@{}}
         \ma{\Theta}_{00} & \ma{0} \\\hline
         \ma{0} & \ma{0}
    \end{array}\right] +
    \left[\begin{array}{@{}c|c@{}}
         \ma{0} & \ma{\Theta}_{0k} \\\hline
         \ma{\Theta}_{0k}^\top & \ma{\Theta}_{kk}
    \end{array}\right] \\
    &=  \left[
    \begin{array}{@{}c|c@{}}
         \ma{\Theta}_{00} & \ma{\Theta}_{0k} \\\hline
         \ma{\Theta}_{0k}^\top & \ma{\Theta}_{kk}
    \end{array}\right]
    \\
    &= \left[
    \begin{array}{@{}ccc|c@{}}
        a_{11}\ma{I} + r_{11} \ma{11}^\top & r_{12} \ma{11}^\top & \cdots & r_{13} \ma{11}^\top \\
        r_{21} \ma{11}^\top & a_{22}\ma{I} + r_{22} \ma{11}^\top & \cdots & r_{2k} \ma{11}^\top \\
        \vdots & \vdots & \ddots & \vdots \\\hline
        r_{k1} \ma{11}^\top & r_{k2} \ma{11}^\top & \cdots & a_{kk}\ma{I} + r_{kk} \ma{11}^\top \\
    \end{array}\right]
\end{align*}
Consequently, updating all parameters in $\ma{A}$ and $\ma{R}$ that are associated with cluster $k$ updates an entire block of variables in $\ma{\Theta}$.
To separate the term $-\log|\ma{P}^{1/2} \ma{R}^* \ma{P}^{1/2}|$ in objective function \eqref{eq:loss-decomposed}, it helps to write $\ma{P}^{1/2} \ma{R}^* \ma{P}^{1/2}$ as the $2 \times 2$ block matrix
\begin{equation*}
    \ma{P}^{1/2}\ma{R}^*\ma{P}^{1/2} =
    \left[
    \begin{array}{@{}c|c@{}}
    \ma{P}_{0}^{1/2}\ma{R}_{0}^*\ma{P}_{0}^{1/2}  & p_k^{1/2}\ma{P}_{0}^{1/2}\ma{r}_{k}   \\ \hline
        p_k^{1/2}(\ma{P}_{0}^{1/2}\ma{r}_{k})^\top & p_k r_{kk}^* \\
    \end{array}
    \right]
\end{equation*}
with $\ma{r}_{k}$ the $K-1$ vector containing all elements in the $k^{\text{th}}$ column of $\ma{R}$ apart from $r_{kk}$, and
\begin{equation*}
    \ma{P} =
    \left[
    \begin{array}{@{}cc@{}}
        \ma{P}_{0} & \ma{0}  \\
        \ma{0}^\top & p_k \\
    \end{array}
    \right].
\end{equation*}
From linear algebra, we have the following expression for the log determinant of the $2 \times 2$ block matrix
\begin{equation}
\label{app:eq:determinant_decomposition_k}
    -\log|\ma{P}^{1/2} \ma{R}^* \ma{P}^{1/2}| = -\log |\ma{P}_{0}^{1/2}\ma{R}_{0}^*\ma{P}_{0}^{1/2}|
    - \log(a_{kk} + p_k r_{kk} - p_k \ma{r}_{k}^\top (\ma{R}_{0}^*)^{-1}\ma{r}_{k}).
\end{equation}
This allows us to rewrite the negative log determinant of $\ma{\Theta}$ as
\begin{align}
    -\log | \ma{\Theta} | ={}& -\log |\ma{P}_{0}^{1/2}\ma{R}_{0}^*\ma{P}_{0}^{1/2}|
    - \log(a_{kk} + p_k r_{kk} - p_k \ma{r}_{k}^\top (\ma{R}_{0}^*)^{-1}\ma{r}_{k}) \nonumber\\
    &- (p_k - 1)\log a_{kk} - \sum_{\substack{\ell=1 \\\ell \neq k}}^K (p_\ell - 1)\log a_{\ell\ell}. \nonumber
\end{align}
Using the re-expression from objective function \eqref{eq:loss-decomposed}, the trace of $\ma{S}\ma{\Theta}$ separates readily into
\begin{equation*}
  \tr \ma{S\Theta} = 2 \ma{u}_k^\top \ma{S} \ma{U}_0 \ma{r}_k + \ma{u}_k^\top \ma{S} \ma{u}_k r_{kk} + a_{kk} \tr \ma{S}_k + \tr\ma{S} \ma{U}_0 \ma{R}_0 \ma{U}_0^\top + \sum_{\substack{\ell=1\\\ell \neq k}}^K a_{\ell\ell} \tr \ma{S}_\ell.
\end{equation*}
It can be verified that the last two terms are constant in $a_{kk}$, $\ma{r}_k$, and $r_{kk}$, and the remaining terms are linear in these unknowns.
Additionally, the sparsity penalty can be written as
\begin{equation*}
    \sum_{k=1}^K \sum_{\ell=1}^K \ma{u}_k \ma{Z} \ma{u}_{\ell}^\top  |r_{k\ell}| = \sum_{\ell=1}^K \ma{u}_k \ma{Z} \ma{u}_{\ell}^\top  |r_{k\ell}| + \sum_{\ell,\ell' \neq k} \ma{u}_{\ell} \ma{Z} \ma{u}_{\ell'}^\top  |r_{\ell\ell'}|.
\end{equation*}
Finally, due to the symmetry of $\ma{R}$, the clusterpath penalty is not separable into parts dependent and independent of cluster $k$.

Combining the foregoing results,
objective function $L(\ma{\Theta}_k)$
can be expressed in terms of parts $a_{kk}$, $\ma{r}_k$, and $r_{kk}$, and we obtain
\begin{align*}
    L(a_{kk}, \ma{r}_k, r_{kk}) ={}& L_{\text{det}}(a_{kk}, \ma{r}_k, r_{kk}) + L_{\text{cov}}(a_{kk}, \ma{r}_k, r_{kk})  +  L_{\text{clust}}(a_{kk}, \ma{r}_k, r_{kk}) \\
    &+ L_{\text{sparse}}(\ma{r}_k, r_{kk}) + C \\
    ={}& - \log(a_{kk} + p_k r_{kk} - p_k \ma{r}_{k}^\top (\ma{R}_{0}^*)^{-1}\ma{r}_{k})
    -(p_k - 1)\log a_{kk} \\
    & + 2 \ma{u}_k^\top \ma{S} \ma{U}_0 \ma{r}_k + \ma{u}_k^\top \ma{S} \ma{u}_k r_{kk} + a_{kk} \tr \ma{S}_k  \\
    & + \lambda_c {\sum_{k=1}^K \sum_{\ell=1}^{k-1} \ma{u}_k^\top \ma{W} \ma{u}_\ell d_{\mathcal{C}_k \mathcal{C}_\ell}(\ma{A}, \ma{R})} + \lambda_s \sum_{\ell=1}^K \ma{u}_k \ma{Z} \ma{u}_{\ell}^\top  |r_{k\ell}| +~C
\end{align*}
with
\begin{align*}
    C ={}& -\log |\ma{P}_{0}^{1/2}\ma{R}_{0}^*\ma{P}_{0}^{1/2}| - \sum_{\substack{\ell=1 \\\ell \neq k}}^K (p_\ell - 1)\log a_{\ell\ell} \\
    &+ \tr\ma{S} \ma{U}_0 \ma{R}_0 \ma{U}_0^\top + \sum_{\substack{\ell=1\\\ell \neq k}}^K a_{\ell\ell} \tr \ma{S}_\ell + \lambda_s \sum_{\ell,\ell' \neq k} \ma{u}_{\ell} \ma{Z} \ma{u}_{\ell'}^\top  |r_{\ell\ell'}|.
\end{align*}

\section{Derivations and Pseudocode for the Cyclic Block Coordinate Descent Algorithm} \label{app:alogirthm-details}
We provide a detailed explanation of the minimization procedure for the CGGM objective function. In Appendix~\ref{appendix:algorithm}, we present the pseudocode of the algorithm. We derive the gradient and Hessian used in the algorithm in Appendix~\ref{app:subsec:gradientandhessian}. Finally, Appendix~\ref{app:subsec:alg_details} discusses the computation of a clusterpath, where the CGGM objective is minimized for a sequence of increasing values of the tuning parameter $\lambda_c$.

The reparameterization of $\ma{\Theta}$ into
the diagonal matrix $\ma{A}$ and the symmetric matrix $\ma{R}$ (see Appendix~\ref{app:clusterpathderivatation}) allows for
straightforward expressions of the objective function and the conditions for positive definiteness of $\ma{\Theta}$. In this appendix, we take an additional step to facilitate the derivations of the gradient and Hessian used for the minimization.

We define $a_{kk} = b_{kk} - r_{kk}$, so that the diagonal values of $\ma{\Theta}$ are solely determined by the diagonal matrix $\ma{B} = (b_{kk})_{1\leq k\leq K}$. This
allows for more compact representation of the gradient and Hessian with respect to $r_{kk}$. We work with the following definition of the objective function
\begin{align}
    L(b_{kk}, \ma{r}_k, r_{kk}) ={}& - \log(b_{kk} + (p_k - 1) r_{kk} - p_k \ma{r}_{k}^\top (\ma{R}_{0}^*)^{-1}\ma{r}_{k})
    -(p_k - 1)\log (b_{kk} - r_{kk}) \nonumber\\
    & + 2 \ma{u}_k^\top \ma{S} \ma{U}_0 \ma{r}_k + \ma{u}_k^\top \ma{S} \ma{u}_k r_{kk} + (b_{kk} - r_{kk}) \tr \ma{S}_k  \nonumber\\
    & + \lambda_c {\sum_{k=1}^K \sum_{\ell=1}^{k-1} \ma{u}_k^\top \ma{W} \ma{u}_\ell d_{\mathcal{C}_k \mathcal{C}_\ell}(\ma{B}, \ma{R})} + \lambda_s \sum_{\ell=1}^K \ma{u}_k \ma{Z} \ma{u}_{\ell}^\top  |r_{k\ell}| +~C \label{app:eq:loss_with_B}
\end{align}
with
\begin{equation*}
    d_{\mathcal{C}_k \mathcal{C}_\ell}^2(\ma{B},\ma{R}) = (b_{kk} - b_{\ell\ell})^2 + (p_k - 1)(r_{kk} - r_{k\ell})^2 + (p_\ell - 1)(r_{\ell\ell} - r_{k\ell})^2 + \sum_{\substack{m=1\\  m \notin \{k, \ell\}}}^K p_m(r_{km} - r_{\ell m})^2.
\end{equation*}

\subsection{Pseudocode for the Algorithm} \label{appendix:algorithm}

Algorithm~\ref{alg:cggm} contains the pseudocode for the cyclic block coordinate descent algorithm to compute CGGM.

\begin{algorithm}[p]
\setstretch{1.35}
\caption{Pseudocode for the algorithm that minimizes the CGGM objective function for fixed $\lambda_c$ and $\lambda_s$}
\label{alg:cggm}
\hspace*{\algorithmicindent} \textbf{Input} Initial estimates for $\ma{B}^{(1)}$ and $\ma{R}^{(1)}$, sample covariance matrix \ma{S}, weight matrix \hspace*{\algorithmicindent} \ma{W}, tuning parameters $\lambda_c$ and $\lambda_s$, thresholds $\varepsilon_f$ and $\varepsilon_s$, convergence threshold $\varepsilon_c$, maxi-  \hspace*{\algorithmicindent} mum number of iterations $t_\text{max}$ \\
\hspace*{\algorithmicindent} \textbf{Output} $\widehat{\ma{\Theta}}$ that minimizes $L(\ma{\Theta})$ and $\hat{K}$ the number of clusters
\begin{algorithmic}[1]
\State $\ma{U}^{(1)} \leftarrow \ma{I}$
\State $L^{(1)} \leftarrow L(\ma{B}^{(1)}, \ma{R}^{(1)})$
\State $L^{(0)} \leftarrow (1 + 2 \varepsilon_c)L^{(1)}$
\State $K \leftarrow p$
\State $t \leftarrow 1$
\While{$L^{(t-1)} / L^{(t)} - 1 > \varepsilon_c$ and $t \leq t_\text{max}$}
    \State $t \leftarrow t + 1$
    \State $\ma{B}^{(t)} \leftarrow \ma{B}^{(t-1)}$
    \State $\ma{R}^{(t)} \leftarrow \ma{R}^{(t-1)}$
    \State $\ma{U}^{(t)} \leftarrow \ma{U}^{(t-1)}$
    \For{$k=1,\ldots,K$}
        \State $\ell \leftarrow \argmin_{\ell'} d_{k\ell'} (\ma{B}^{(t)}, \ma{R}^{(t)})$
        \If{$d_{\mathcal{C}_k \mathcal{C}_\ell}(\ma{B}^{(t)}, \ma{R}^{(t)}) \leq \varepsilon_f$}
            \State Fuse clusters $k$ and $\ell$ by modifying $\ma{B}^{(t)}$, $\ma{R}^{(t)}$, $\ma{U}^{(t)}$
            \State $K \leftarrow K - 1$
        \Else
            \State $\bm\delta_k \leftarrow -\nabla^2 L \bigl(b^{(t)}_{kk}, \ma{r}^{(t)}_k, r^{(t)}_{kk} \bigr) ^{-1} \nabla L \bigl(b^{(t)}_{kk}, \ma{r}^{(t)}_k, r^{(t)}_{kk} \bigr) $
            \State Compute maximum step size $s_\text{max}$ using equations \eqref{eq:s_max1} and \eqref{eq:s_max2}
            \State Select optimal step size $s^* \in [0, s_\text{max})$
            \State $ \bigl[b^{(t)}_{kk}, \ma{r}^{(t)}_k, r^{(t)}_{kk} \bigr] \leftarrow \bigl[b^{(t)}_{kk}, \ma{r}^{(t)}_k, r^{(t)}_{kk} \bigr] + s^*\bm\delta_k$
        \EndIf
    \EndFor
\EndWhile
\State $\ma{r}^{(t)}_{k\ell} \leftarrow 0$ if $|\ma{r}^{(t)}_{k\ell}| < \varepsilon_s$ for all $(k, \ell)$
\State $\widehat{\ma{B}} \leftarrow \ma{B}^{(t)}$
\State $\widehat{\ma{R}} \leftarrow \ma{R}^{(t)}$
\State $\hat{K} \leftarrow K$
\State $\widehat{\ma{A}} \leftarrow \text{diag}(\hat{a}_{11} \ma{I}, \ldots, \hat{a}_{KK} \ma{I})$ where $\hat{a}_{kk} = \hat{b}_{kk} - \hat{r}_{kk}$
\State $\widehat{\ma{U}} \leftarrow \ma{U}^{(t)}$
\State $\widehat{\ma{\Theta}} \leftarrow \widehat{\ma{U}} \widehat{\ma{R}} \widehat{\ma{U}}^\top + \widehat{\ma{A}}$
\end{algorithmic}
\end{algorithm}

The algorithm requires initialization; the usage of high-quality initializations is important for the stability of the algorithm in terms of convergence.
 To initialize $\ma{B}$ and $\ma{R}$ for $\lambda_c=\lambda_s=0$, one can use the inverse of the sample covariance matrix if it is available; then $\ma{B}^{(1)} = \text{diag}(\ma{S}^{-1})$ and $\ma{R}^{(1)} = \ma{S}^{-1}$.
In other cases, one can start from $(\ma{S} + \ma{I})^{-1}$ or a regularized estimator such as the graphical lasso (e.g., \citealp{friedman2008glasso, witten2011new}).
Next, when computing the clusterpath, namely a sequence of solutions for $\mathbf{\Theta}$ for an increasing clustering parameter $\lambda_c$ (for fixed $\lambda_s$), we leverage  warm starts to ensure high-quality initializations. Indeed, after solving the minimization problem for a particular $\lambda_c$, we use that solution as initialization when solving the optimization problem for the next (higher) value of $\lambda_c$ in the clusterpath.  Loss function convergence of the algorithm was observed across all numerical experiments and empirical applications.

After the required initializations, the algorithm first checks for eligible clusters fusions using the fusion threshold $\varepsilon_f$ (lines~12--15). The value of $\varepsilon_f$ can be a small user-defined value, such as $10^{-3}$, or based on the data itself. If the inverse of the sample covariance matrix is available, a data-driven choice is $\varepsilon_f = \tau \median_{j,j'} (d_{jj'}(\ma{S}^{-1}))$, with $\tau=10^{-3}$.
If a fusion is performed, the optimization parameters are modified and the total number of clusters $K$ is reduced by one. In case no cluster fusion occurs, the algorithm proceeds to the second step where the parameters pertaining to cluster $k$, denoted by the vector $[b_{kk}, \ma{r}_k, r_{kk}]$, are jointly updated (lines~17--20). Part of the update is a line search by means of a golden section search for the optimal step size $s^*$, which ensures the positive definiteness of $\ma{\Theta}$.

As mentioned in Appendix~\ref{app:clusterpathderivatation}, $\ma{\Theta}$ is positive definite if $\ma{R}^*$ is positive definite and $a_{kk}=b_{kk}+r_{kk}>0$. Assume that $\ma{R}^*$ is positive definite for the current values of $[b_{kk}, \ma{r}_k, r_{kk}]$. Then, Sylvester's Criterion states that the upper left $1 \times 1$ corner of $\ma{R}^*$ has a positive determinant, so does the $2 \times 2$ upper left corner, and so on until $\ma{R}^*$ itself has a positive determinant. Without loss of generality, let the clusters be arranged so that $k$ is the last cluster. Updating $[b_{kk}, \ma{r}_k, r_{kk}]$ does not alter the $1 \times 1$ through $(K-1) \times (K-1)$ upper left corners. Consequently, a positive determinant of $\ma{R}^*$ after the update of the parameters related to cluster $k$ is sufficient to keep $\ma{R}^*$ positive definite.

Putting this together, the positive definiteness of $\ma{\Theta}$ is preserved by a step size $s$ that satisfies the aforementioned conditions. Using the decomposition of the log determinant in \eqref{app:eq:determinant_decomposition_k} and the expression on the first line of the objective in \eqref{app:eq:loss_with_B}, this is achieved by a step size that satisfies the inequalities
\begin{align}
\label{eq:s_max1}
    (b_{kk} + s \delta_{b_{kk}}) + (p_k - 1) (r_{kk} + s \delta_{r_{kk}}) - p_k (\ma{r}_k + s \bm\delta_{\ma{r}_k})^\top (\ma{R}^{*})^{-1} (\ma{r}_k + s \bm\delta_{\ma{r}_k}) &> 0 \nonumber\\
    b_{kk} + s \delta_{b_{kk}} - r_{kk} - s \delta_{r_{kk}} &> 0,
\end{align}
where $\delta_{b_{kk}}$, $\bm\delta_{\ma{r}_k}$, and $\delta_{r_{kk}}$ represent the descent directions for $b_{kk}$, $\ma{r}_k$, and $r_{kk}$.
In case there is a single cluster left ($K=1$), the first inequality reduces to
\begin{equation}
\label{eq:s_max2}
    (b_{kk} + s \delta_{b_{kk}}) + (p_k - 1) (r_{kk} + s \delta_{r_{kk}}) > 0.
\end{equation}
After the loss function has converged, a thresholding operation is applied to $\ma{R}$ to set sufficiently small elements to zero and induce sparsity. Finally,
the output of the algorithm consists of the estimated number of clusters $\hat{K}$ and the estimates $\widehat{\ma{B}}$, $\widehat{\ma{R}}$, and $\widehat{\ma{U}}$, which can be used to construct $\widehat{\ma{\Theta}}$.

The computational complexity of the algorithm is primarily determined by the computation of the descent direction $\bm{\delta}_k$, which requires solving a system of equations with a complexity of $\mathcal{O}(K^3)$. This step dominates the overall update cost, including the line search. To determine $s^*$, we use the golden section search, which iteratively shrinks the interval $[0, s_{\max}]$ to a pre-specified tolerance of $5 \cdot 10^{-3}$. During this search, the objective for cluster $k$ in \eqref{app:eq:loss_with_B} is repeatedly evaluated, incurring a per-evaluation complexity of $\mathcal{O}(K^2)$. Consequently, a complete pass over all clusters results in a total complexity of $\mathcal{O}(K^4)$. Since the number of clusters $K$ is bounded by the number of variables $p$, the overall complexity is capped at $\mathcal{O}(p^4)$.

\subsection{Derivations of the Gradient and Hessian}
\label{app:subsec:gradientandhessian}
For simplicity, we use the following smoothed version of the absolute value function for the sparsity penalty
\begin{equation*}
    |r_{k\ell}| =
    \begin{cases}
        \frac{r_{k\ell}^2 + \varepsilon_s^2}{2 \varepsilon_{s}} & \text{if $|r_{k\ell}| < \varepsilon_{s}$,} \\
        |r_{k\ell}| & \text{otherwise}.
    \end{cases}
\end{equation*}

If $\varepsilon_{s}$ is chosen sufficiently small (e.g., $\varepsilon_{s}=5 \cdot 10^{-3}$), this function closely approximates the absolute value function.

To shorten notation, let $\ma{V} = (\ma{R}_{0}^*)^{-1}$ and
\begin{equation*}
    h(b_{kk}, \ma{r}_k, r_{kk}) = b_{kk} + (p_k - 1) r_{kk} - p_k \ma{r}_{k}^\top \ma{V} \ma{r}_{k}.
\end{equation*}
For convenience, the elements $r_{km}$ of the vector $\ma{r}_k$ are treated individually. Consequently, we should note that $1 \leq m \leq K$, $m \neq k$. For the gradient, we obtain
\begin{align*}
    \fracpartial{L_{\text{det}}(b_{kk}, \ma{r}_k, r_{kk})}{b_{kk}} ={}& -\frac{1}{h(b_{kk}, \ma{r}_k, r_{kk})} - \frac{p_k - 1}{b_{kk} - r_{kk}}\\
    \fracpartial{L_{\text{det}}(b_{kk}, \ma{r}_k, r_{kk})}{r_{km}} ={}& \frac{2 p_k}{h(b_{kk}, \ma{r}_k, r_{kk})} \ma{v}_m^\top \ma{r}_k\\
    \fracpartial{L_{\text{det}}(b_{kk}, \ma{r}_k, r_{kk})}{r_{kk}} ={}& -\frac{p_k - 1}{h(b_{kk}, \ma{r}_k, r_{kk})} + \frac{p_k - 1}{b_{kk} - r_{kk}}\\
    \fracpartial{L_{\text{cov}}b_{kk}, \ma{r}_k, r_{kk})}{b_{kk}} ={}& \tr \ma{S}_k \\
    \fracpartial{L_{\text{cov}}(b_{kk}, \ma{r}_k, r_{kk})}{r_{km}} ={}& 2 \ma{u}_m^\top \ma{S} \ma{u}_k\\
    \fracpartial{L_{\text{cov}}(b_{kk}, \ma{r}_k, r_{kk})}{r_{kk}} ={}& \ma{u}_k^\top \ma{S} \ma{u}_k - \tr \ma{S}_k \\
    \fracpartial{L_{\text{clust}}(b_{kk}, \ma{r}_k, r_{kk})}{b_{kk}} ={}& \lambda \sum_{\substack{\ell = 1 \\ \ell \neq k}}^K  \frac{\ma{u}_k^\top \ma{W} \ma{u}_\ell}{d_{\mathcal{C}_k \mathcal{C}_\ell}(\ma{B}, \ma{R})} (b_{kk} - b_{\ell\ell}) \\
    \fracpartial{L_{\text{clust}}(b_{kk}, \ma{r}_k, r_{kk})}{r_{km}} ={}& \lambda \sum_{\substack{\ell=1 \\ \ell \notin \{ k, m \}}}^K \left(\frac{\ma{u}_k^\top \ma{W} \ma{u}_\ell}{d_{\mathcal{C}_k \mathcal{C}_\ell}(\ma{B}, \ma{R})} p_m (r_{km} - r_{m\ell}) + \frac{\ma{u}_m^\top \ma{W} \ma{u}_\ell}{d_{\mathcal{C}_m \mathcal{C}_\ell}(\ma{B}, \ma{R})} p_k (r_{km} - r_{k\ell})\right) \\
    &+ \lambda \frac{\ma{u}_k^\top \ma{W} \ma{u}_m}{d_{\mathcal{C}_k \mathcal{C}_m}(\ma{B}, \ma{R})}((p_k - 1)(r_{km} - r_{kk}) + (p_m - 1)(r_{km} - r_{mm})) \\
    \fracpartial{L_{\text{clust}}(b_{kk}, \ma{r}_k, r_{kk})}{r_{kk}} ={}& \lambda (p_k - 1) \sum_{\substack{\ell = 1 \\ \ell \neq k}}^K \frac{\ma{u}_k^\top \ma{W} \ma{u}_\ell}{d_{\mathcal{C}_k \mathcal{C}_\ell}(\ma{B}, \ma{R})} (r_{kk} - r_{k\ell}) \\
    \fracpartial{L_{\text{sparse}}(b_{kk}, \ma{r}_k, r_{kk})}{r_{k\ell}} ={}&
    \begin{cases}
        \frac{\ma{u}_k \ma{Z} \ma{u}_{\ell}^\top}{\varepsilon_{s}} r_{k\ell} & \text{if $|r_{k\ell}| < \varepsilon_{s}$,} \\
        \text{sign}(r_{k\ell}) & \text{otherwise}.
    \end{cases}
\end{align*}
In the derivations for the Hessian, we require an additional index $m'$ to define the off-diagonal elements in the block $\partial^2 L(b_{kk}, \ma{r}_k, r_{kk})/\partial \ma{r}_k^2$ that satisfies $1 \leq m' \leq K$, $m' \notin \{ k, m \}$. For the Hessian, we obtain
\begin{align*}
    \frac{\partial^2 L_{\text{det}}(b_{kk}, \ma{r}_k, r_{kk})}{\partial b_{kk}^2} ={}& \frac{1}{h^2(b_{kk}, \ma{r}_k, r_{kk})} + \frac{p_k - 1}{(b_{kk} - r_{kk})^2} \\
    \frac{\partial^2 L_{\text{det}}(b_{kk}, \ma{r}_k, r_{kk})}{\partial b_{kk}\partial r_{km}} ={}& -\frac{2 p_k}{h^2(b_{kk}, \ma{r}_k, r_{kk})} \ma{v}_m^\top \ma{r}_k \\
    \frac{\partial^2 L_{\text{det}}(b_{kk}, \ma{r}_k, r_{kk})}{\partial b_{kk} \partial r_{kk}} ={}& \frac{p_k - 1}{h^2(b_{kk}, \ma{r}_k, r_{kk})} - \frac{p_k - 1}{(b_{kk} - r_{kk})^2} \\
    \frac{\partial^2 L_{\text{det}}(b_{kk}, \ma{r}_k, r_{km})}{\partial r_{km}^2} ={}& \frac{2p_k}{h(b_{kk}, \ma{r}_k, r_{kk})} v_{mm} + \frac{4 p_k^2}{h^2(b_{kk}, \ma{r}_k, r_{kk})} \ma{v}_m^\top \ma{r}_k \ma{r}_k^\top \ma{v}_{m} \\
    \frac{\partial^2 L_{\text{det}}(b_{kk}, \ma{r}_k, r_{km})}{\partial r_{km} \partial r_{km'}} ={}& \frac{2p_k}{h(b_{kk}, \ma{r}_k, r_{kk})} v_{mm'} + \frac{4 p_k^2}{h^2(b_{kk}, \ma{r}_k, r_{kk})} \ma{v}_m^\top \ma{r}_k \ma{r}_k^\top \ma{v}_{m'} \\
    \frac{\partial^2 L_{\text{det}}(b_{kk}, \ma{r}_k, r_{kk})}{\partial r_{km} \partial r_{kk}} ={}& -\frac{2 p_k (p_k - 1)}{h^2(b_{kk}, \ma{r}_k, r_{kk})} \ma{v}_m^\top \ma{r}_k \\
    \frac{\partial^2 L_{\text{det}}(b_{kk}, \ma{r}_k, r_{kk})}{\partial r_{kk}^2} ={}& \frac{(p_k - 1)^2}{h^2(b_{kk}, \ma{r}_k, r_{kk})} + \frac{p_k - 1}{(b_{kk} - r_{kk})^2} \\
    \frac{\partial^2 L_{\text{clust}}(b_{kk}, \ma{r}_k, r_{kk})}{\partial b_{kk}^2} ={}& \lambda \sum_{\substack{\ell = 1 \\ \ell \neq k}}^K  \ma{u}_k^\top \ma{W} \ma{u}_\ell \left( \frac{1}{d_{\mathcal{C}_k \mathcal{C}_\ell}(\ma{B}, \ma{R})} - \frac{(b_{kk} - b_{\ell\ell})^2}{d_{\mathcal{C}_k \mathcal{C}_\ell}^3(\ma{B}, \ma{R})} \right) \\
    \frac{\partial^2 L_{\text{clust}}(b_{kk}, \ma{r}_k, r_{kk})}{\partial b_{kk} \partial r_{km}} ={}& -\lambda \sum_{\substack{\ell=1 \\ \ell \notin \{ k, m \}}}^K \frac{\ma{u}_k^\top \ma{W} \ma{u}_\ell}{d_{\mathcal{C}_k \mathcal{C}_\ell}^3(\ma{B}, \ma{R})} p_m (b_{kk} - b_{\ell\ell}) (r_{km} - r_{m\ell}) \\
    &- \lambda \frac{\ma{u}_k^\top \ma{W} \ma{u}_m}{d_{\mathcal{C}_k \mathcal{C}_m}^3(\ma{B}, \ma{R})} (b_{kk} - b_{mm}) \Bigl((p_k - 1)(r_{km} - r_{kk}) \\
    &\pushright{+\,(p_m - 1)(r_{km} - r_{mm})\Bigr)} \\
    \frac{\partial^2 L_{\text{clust}}(b_{kk},\ma{r}_k, r_{kk})}{\partial b_{kk} \partial r_{kk}} ={}& -\lambda (p_k - 1) \sum_{\substack{\ell = 1 \\ \ell \neq k}}^K \frac{\ma{u}_k^\top \ma{W} \ma{u}_\ell}{d_{\mathcal{C}_k \mathcal{C}_\ell}^3(\ma{B}, \ma{R})} (b_{kk} - b_{\ell\ell})(r_{kk} - r_{k\ell}) \\
    \frac{\partial^2 L_{\text{clust}}(b_{kk}, \ma{r}_k, r_{kk})}{\partial r_{km}^2} ={}& \lambda \sum_{\substack{\ell=1 \\ \ell \notin \{ k, m \}}}^K p_m \frac{\ma{u}_k^\top \ma{W} \ma{u}_\ell}{d_{\mathcal{C}_k \mathcal{C}_\ell}(\ma{B}, \ma{R})} \left(1 -\frac{p_m (r_{km} - r_{m\ell})^2}{d_{\mathcal{C}_k \mathcal{C}_\ell}^2(\ma{B}, \ma{R})} \right) \\
    &+ \lambda \sum_{\substack{\ell=1 \\ \ell \notin \{ k, m \}}}^K p_k \frac{\ma{u}_m^\top \ma{W} \ma{u}_\ell}{d_{\mathcal{C}_m \mathcal{C}_\ell}(\ma{B}, \ma{R})} \left( 1 -  \frac{p_k (r_{km} - r_{k\ell})^2}{d_{\mathcal{C}_m \mathcal{C}_\ell}^2(\ma{B}, \ma{R})} \right) \\
    &+ \lambda \frac{\ma{u}_k^\top \ma{W} \ma{u}_m}{d_{\mathcal{C}_k \mathcal{C}_m}(\ma{B}, \ma{R})} \biggl(p_k + p_m - 2 \\
    & \pushright{-\,\frac{((p_k - 1)(r_{km} - r_{kk}) + (p_m - 1)(r_{km} - r_{mm}))^2}{d_{\mathcal{C}_k \mathcal{C}_m}^2(\ma{B}, \ma{R})} \biggr)} \\
    \frac{\partial^2 L_{\text{clust}}(b_{kk}, \ma{r}_k, r_{kk})}{\partial r_{km} \partial r_{km'}} ={}& -\lambda \sum_{\substack{\ell=1 \\ \ell \notin \{ k, m, m' \}}}^K \frac{\ma{u}_k^\top \ma{W} \ma{u}_\ell}{d_{\mathcal{C}_k \mathcal{C}_\ell}^3(\ma{B}, \ma{R})} p_m p_{m'} (r_{km} - r_{m\ell}) (r_{km'} - r_{m'\ell}) \\
    &- \lambda \frac{\ma{u}_k^\top\ma{W}\ma{u}_{m'}}{d_{\mathcal{C}_k \mathcal{C}_{m'}}^3(\ma{B}, \ma{R})} \Bigl((p_k - 1)(r_{km'} - r_{kk}) \\
    &\pushright{+\,(p_m - 1)(r_{km'} - r_{m'm'})\Bigr) p_m (r_{km} - r_{mm'})} \\
    &+ \lambda p_k \frac{\ma{u}_m^\top\ma{W}\ma{u}_{m'}}{d_{\mathcal{C}_m \mathcal{C}_{m'}}(\ma{B}, \ma{R})} \left( 1 - \frac{p_k(r_{km'} - r_{km})^2}{d_{\mathcal{C}_m \mathcal{C}_{m'}}^2(\ma{B}, \ma{R})} \right) \\
    &- \lambda \frac{\ma{u}_k^\top \ma{W} \ma{u}_m}{d_{\mathcal{C}_k \mathcal{C}_m}(\ma{B}, \ma{R})} p_{m'} (r_{km'} - r_{mm'}) \Bigl((p_k - 1)(r_{km} - r_{kk})\\
    &\pushright{+\,(p_m - 1)(r_{km} - r_{mm})\Bigr)} \\
    \frac{\partial^2 L_{\text{clust}}(b_{kk}, \ma{r}_k, r_{kk})}{\partial r_{km} \partial r_{kk}} ={}&  -\lambda \sum_{\substack{\ell=1 \\ \ell \notin \{ k, m \}}}^K \frac{\ma{u}_k^\top \ma{W} \ma{u}_\ell}{d_{\mathcal{C}_k \mathcal{C}_\ell}^3(\ma{B}, \ma{R})} p_m (p_k - 1) (r_{kk} - r_{k\ell}) (r_{km} - r_{m\ell}) \\
    &- \lambda (p_k - 1) \frac{\ma{u}_k^\top \ma{W} \ma{u}_m}{d_{\mathcal{C}_k \mathcal{C}_m}(\ma{B}, \ma{R})} \left( 1 - \frac{(p_k - 1) (r_{km} - r_{kk})}{d_{\mathcal{C}_k \mathcal{C}_m}^2(\ma{B}, \ma{R})} (r_{km} - r_{kk}) \right. \\
    & \pushright{+ \left. \frac{(p_m - 1) (r_{km} - r_{mm})}{d_{\mathcal{C}_k \mathcal{C}_m}^2(\ma{B}, \ma{R})} (r_{km} - r_{kk}) \right)} \\
    \frac{\partial^2 L_{\text{clust}}(b_{kk}, \ma{r}_k, r_{kk})}{\partial r_{kk}^2} ={}&  \lambda (p_k - 1) \sum_{\substack{\ell = 1 \\ \ell \neq k}}^K \ma{u}_k^\top \ma{W} \ma{u}_\ell \left( \frac{1}{d_{\mathcal{C}_k \mathcal{C}_\ell}(\ma{B}, \ma{R})} - \frac{(p_k - 1) (r_{kk} - r_{k\ell})^2}{d_{\mathcal{C}_k \mathcal{C}_\ell}^3(\ma{B}, \ma{R})} \right) \\
    \frac{\partial^2 L_{\text{sparse}}(b_{kk}, \ma{r}_k, r_{kk})}{\partial r_{k\ell}^2} ={}&
    \begin{cases}
        \frac{\ma{u}_k \ma{Z} \ma{u}_{\ell}^\top}{\varepsilon_{s}} & \text{if $|r_{k\ell}| < \varepsilon_{s}$,} \\
        0 & \text{otherwise}.
    \end{cases}
\end{align*}
where the terms relating to $L_\text{cov}(b_{kk}, \ma{r}_k, r_{kk})$ are left out as they evaluate to zero.
It should be noted that, due to the parameterization of $a_{kk}$ as $b_{kk} - r_{kk}$, if the size of the $k$\textsuperscript{th} cluster is one, the parameter $r_{kk}$ is effectively not a part of the objective function. Consequently, the $k$\textsuperscript{th} element of the gradient is zero and the $k$\textsuperscript{th} row and column of the Hessian are also filled with zeros.

\subsection{Computing  a Clusterpath}
\label{app:subsec:alg_details}
A primary objective of CGGM is to construct a clusterpath, ranging from $p$ to a few clusters. This requires an appropriate sequence for $\lambda_c$, which determines the strength of the aggregation penalty. However, it is not known a priori which value for $\lambda_c$ corresponds to which number of clusters. To aid finding such a sequence, we propose rescaling $\lambda_c$ to reduce its sensitivity to the data and automating the computation of the clusterpath.

First, note that in the current form of objective function \eqref{app:objective_function},
the tuning parameter $\lambda_c$ is sensitive to the number of variables $p$. To reduce this sensitivity, the different terms of the objective function may be scaled. We suggest to scale the terms pertaining to the log likelihood and sparsity penalty by $p^{-1}$ and the clusterpath penalty by \mbox{$\kappa = ((p - 1)^{1/2} \sum_{j<j'} w_{jj'})^{-1}$} to reduce this dependence on $p$. These scaling factors can then easily be absorbed into a rescaled tuning parameter for the clusterpath penalty $\gamma_c = p \kappa \lambda_c$, which then replaces $\lambda_c$ in the objective function.

To set a sequence for $\lambda_c$, we implement an automated procedure. This procedure consists of two stages: a rough stage and a refinement stage. First, the goal is to find the value for $\lambda_c$ as of which the minimum number of clusters is attained. Typically, this is one cluster, but it may be larger if the weight matrix used in the clusterpath penalty is very sparse and contains groups of variables that are not connected through nonzero weights. We initialize $\lambda_c$ at 0.5 and iteratively increment it by 50\% until the minimum number of clusters is reached.
The initial choice $\lambda_c=0.5$ hereby strikes a  balance between the speed at which the maximum value for $\lambda_c$ is found  and the level of detail of the initial clusterpath.
This yields an increasing series of values $\{ \lambda_c^{(q)} : 1 \leq q \leq Q \}$ alongside their corresponding solutions $\{ \widehat{\ma{\Theta}}^{(q)} : 1 \leq q \leq Q \}$.

In the refinement stage, the goal is to obtain a smooth clusterpath. To ensure smoothness in its trajectory, additional values of $\lambda_c$ are inserted whenever the difference between consecutive solutions, as measured by $\| \widehat{\ma{\Theta}}^{(q-1)} - \widehat{\ma{\Theta}}^{(q)} \| / \| \widehat{\ma{\Theta}}^{(q-1)} \|$, exceeds 0.01. Consequently, a continuum of solutions for $\ma{\Theta}$ is obtained, transitioning smoothly from minimal to maximal regularization, with a hierarchical clustering structure. Throughout this iterative process, the algorithm leverages existing solutions for the precision matrix as warm starts to speed up finding solutions for new values of $\lambda_c$.

\newpage
\section{Additional Simulation Results} \label{appendix:simulation}
\begin{figure}[!b]
\centering
\includegraphics[width=0.86\textwidth]{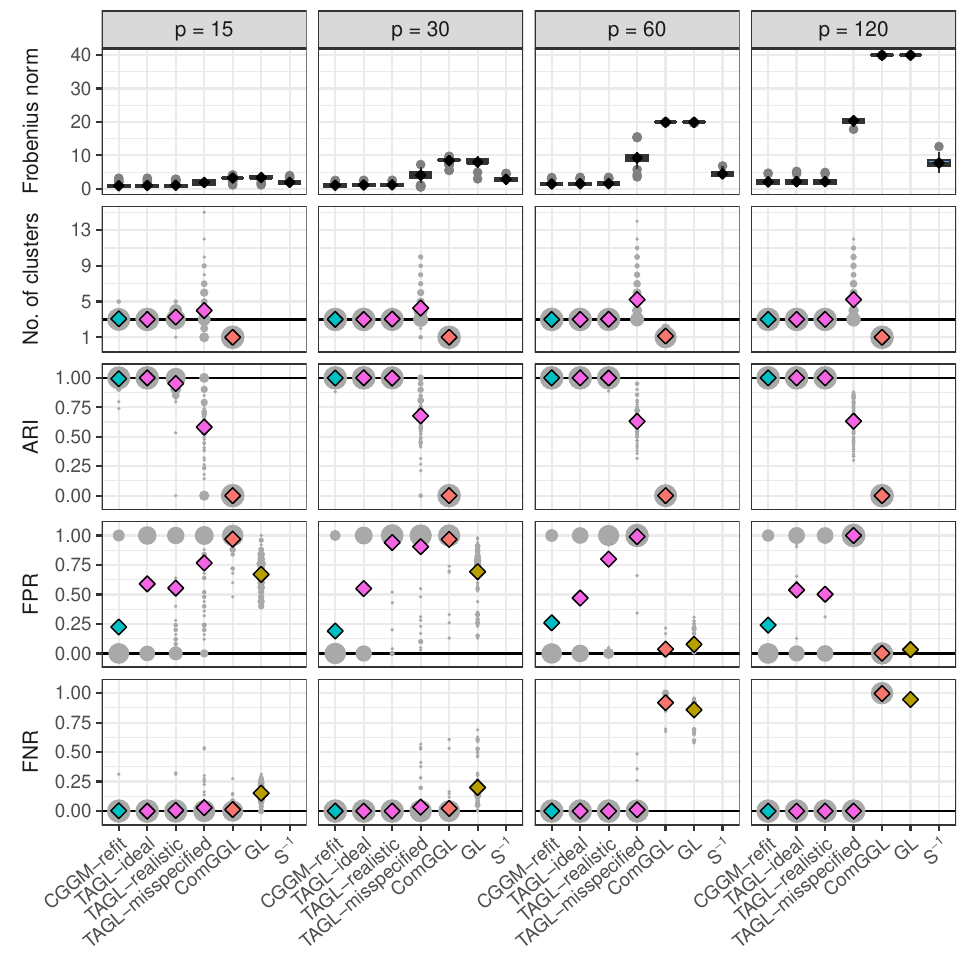}
\vspace{-2pt}
\caption{Results for the chain simulation design with an increasing number of variables (columns). Top row: Boxplots of the Frobenius norm with black diamonds representing the average. Other rows: Diamonds display the average of the estimated number of clusters, ARI, FPR, and FNR. Reference lines are added for the true number of clusters, the ARI value of perfect clustering, and the FPR and FNR of perfect sparsity recognition, respectively. The size of the gray dots represent the frequency of different values across the replications. Aggregation performance is not applicable and omitted for GL and $\ma{S}^{-1}$, as is sparsity recognition performance for $\ma{S}^{-1}$}.
\label{fig:WB2022_variables}
\end{figure}

\begin{figure}[!t]
\centering
\includegraphics[width = 0.86\textwidth]{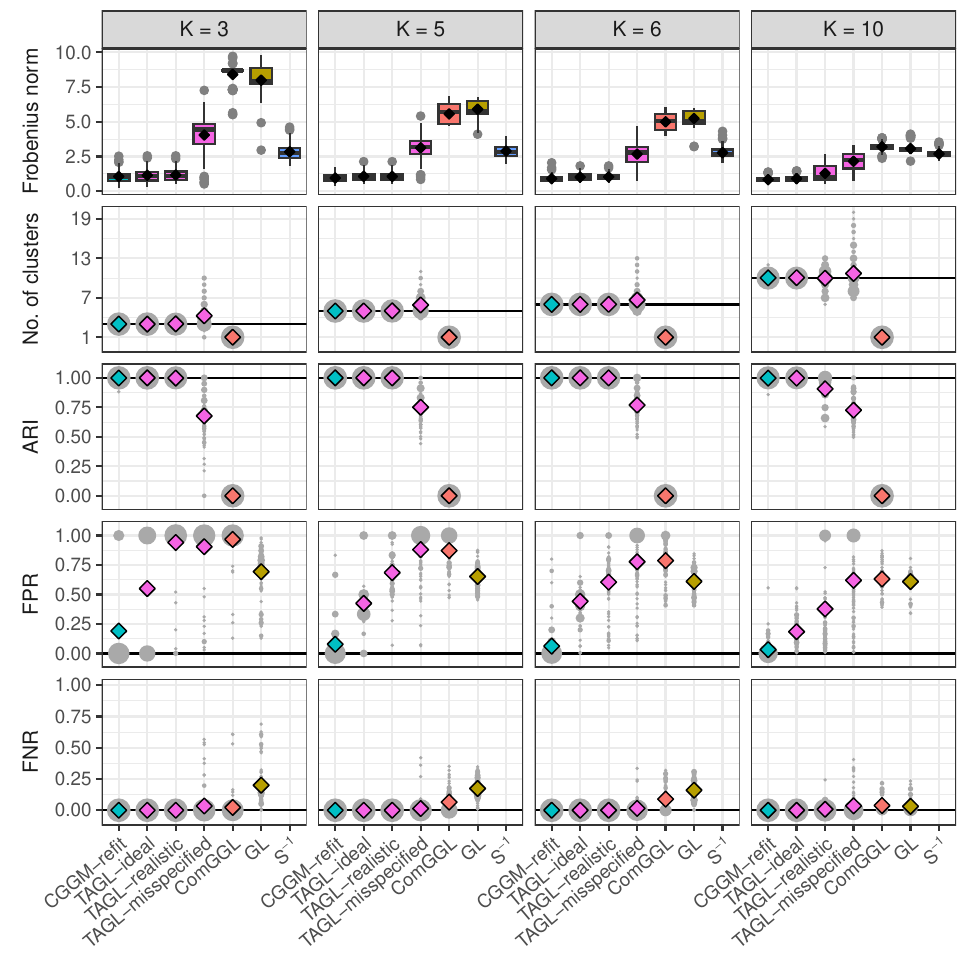}
\vspace{-2pt}
\caption{Results for the chain simulation design with an increasing number of clusters (columns). See Figure \ref{fig:WB2022_variables} for explanatory notes.}
\label{fig:WB2022_clusters}
\end{figure}

\begin{figure}[!t]
\centering
\includegraphics[width = 0.86\textwidth]{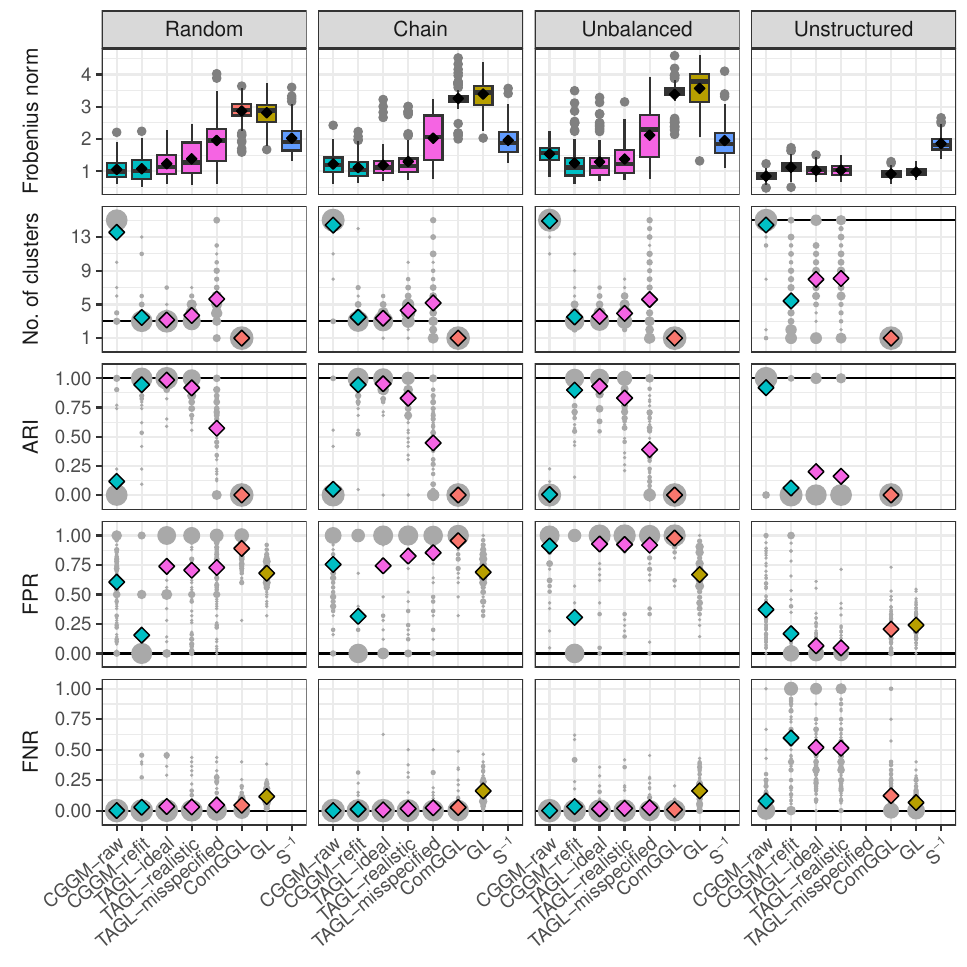}
\vspace{-2pt}
\caption{Results for the simulation designs with approximate block structure (columns). See Figure \ref{fig:WB2022_variables} for explanatory notes. In the unstructured design, a misspecified tree for TAGL does not exist since any tree hierarchy contains the true clustering (each variable being its own cluster).}
\label{fig:imperfect}
\end{figure}

\begin{figure}[!t]
\centering
\includegraphics[width = 0.86\textwidth]{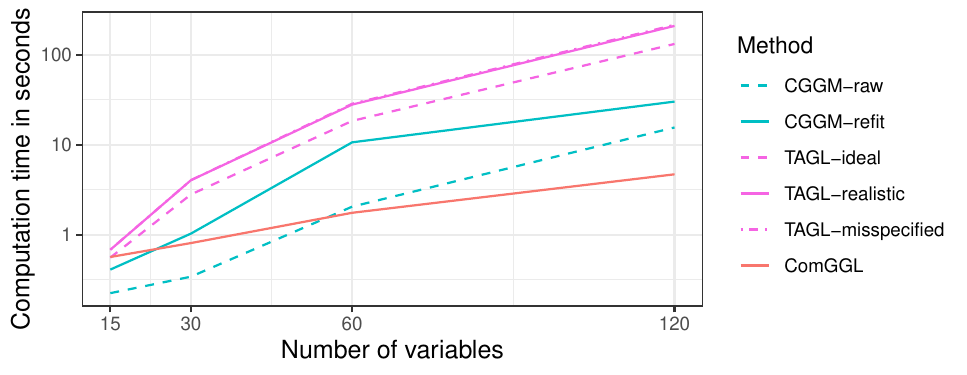}
\caption{Total computation time in seconds across a grid of values for the corresponding aggregation parameter, averaged over 10 replications of the chain simulation design with an increasing number of variables.}
\label{fig:computation_time}
\end{figure}

This appendix provides supplementary simulation results from the experiments discussed in the main paper.
Figures~\ref{fig:WB2022_variables} and~\ref{fig:WB2022_clusters} illustrate the results for the chain simulation design with varying numbers of variables and clusters, respectively.
Figure~\ref{fig:imperfect} shows the results for designs featuring an approximate block structure.
The performance of the methods is evaluated based on estimation accuracy (Frobenius norm), clustering quality (number of clusters and ARI), and sparsity recognition (FPR and FNR). Figure~\ref{fig:computation_time} displays the total computation time over a grid of values for the aggregation parameter (while keeping other tuning parameters fixed) for an increasing number of variables.

\clearpage

\section{Additional Details on Applications} \label{appendix:applications}
We provide additional details on the three empirical applications in Appendices~\ref{appendix:SP100}--\ref{appendix:HSQ}.

\subsection{S\&P 100 Stocks} \label{appendix:SP100}
We collect daily stock price information for the period ranging from January 3\textsuperscript{rd}, 2023, until December 29\textsuperscript{th}, 2023 ($n=250$),
and compute daily realized ranges as given by
\begin{equation*}
    r_{tj} = \frac{(\log H_{tj} - \log L_{tj})^2}{4 \log 2},
\end{equation*}
where $H_{tj}$ and $L_{tj}$ are the high and low prices for stock $j$ during trading day $t$ \citep{parkinson1980extreme} to study the conditional dependency structure of the stocks' realized ranges.

As a preprocessing step, we first fit the popular heterogeneous autoregressive (HAR) model of \cite{corsi2009simple} to the individual daily realized range series to capture time dependencies.
The HAR model captures the temporal dependence in the daily realized range in a very simple yet parsimonious way, namely by explaining it through a weighted average based on the realized range of the preceding day, week and month. After estimating the HAR models, we obtain the $p=101$ standardized residual series and then apply CGGM and TAGL to these to learn the conditional dependency structure among the stocks; our procedure  is in line with \cite{wilms2021tag_lasso}. Note that estimating a graphical model to these residuals series is useful in the context of a multivariate time series analysis to capture the contemporaneous relationships among the $p=101$ realized ranges.

\begin{figure}[!t]
\centering
\includegraphics[width = 0.86\textwidth]{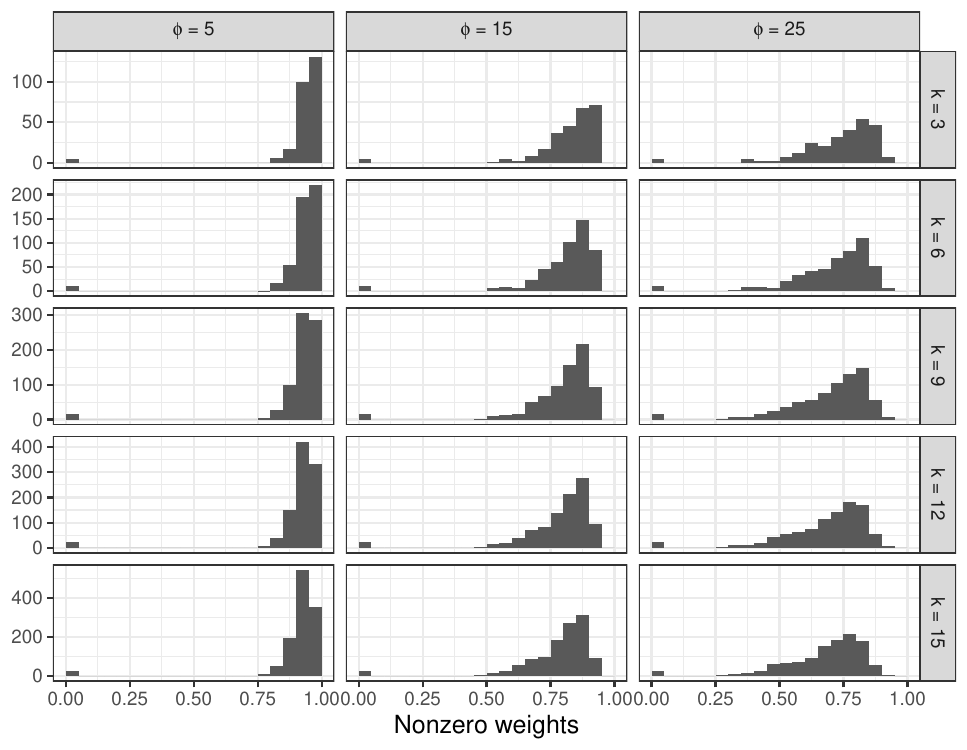}
\caption{Histogram of nonzero clustering weights in CGGM for different candidate values of the tuning parameters $k$ (in rows) and $\phi$ (in columns).}
\label{fig:finance_weights}
\end{figure}

For both CGGM and TAGL, we use 5-fold cross-validation to select the tuning parameters ($k$, $\phi$, $\lambda_{c}$, and $\lambda_{s}$ for CGGM; aggregation and sparsity parameters for TAGL) and refit the precision matrix subject to the obtained variable clustering and sparsity structure. For the clustering weight matrix in CGGM, we use a grid of $k \in \{ 3, 6, 9, 12, 15 \}$ and $\phi \in \{ 5, 15, 25 \}$, based on a visual inspection of the distributions of the resulting nonzero weights (see Figure~\ref{fig:finance_weights}). Clearly, increasing $\phi$ results in a broader distribution of the weights across the interval $(0, 1]$, but all candidate values avoid that the distribution is mostly a singular spike around a specific value. Candidate values for the aggregation and sparsity parameters of CGGM and TAGL are determined via the same procedure as in the simulations from the main text.

\begin{figure}[!t]
\centering
\includegraphics[width = \textwidth]{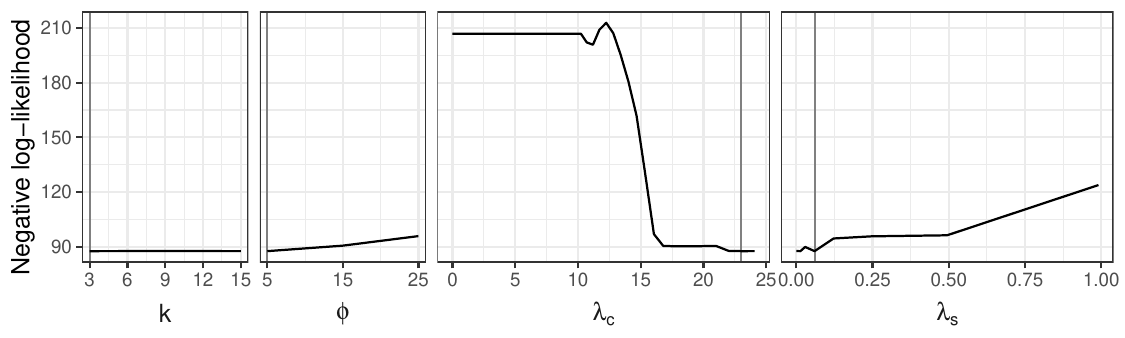}
\caption{Cross-validation score (negative log-likelihood) using the optimal combination of tuning parameters as a baseline. Separate panels visualize the score as one of the tuning parameter varies, with vertical reference lines highlighting the respective optimal values.}
\label{fig:finance_cv}
\end{figure}

To investigate the stability of the tuning parameter selection in CGGM, Figure~\ref{fig:finance_cv} visualizes the cross-validation scores. Since there are four tuning parameters, the optimal combination is used as a baseline (highlighted by vertical reference lines), with separate panels displaying the score as we vary each tuning parameter. Except for some wiggles in the plot for $\lambda_{c}$ and $\lambda_{s}$, the cross-validation score seems to be smooth with respect to the tuning parameters, offering reassurance in the stability of the results.

\subsection{OECD Well-Being Indicators} \label{appendix:OECD}
We analyze the OECD well-being indicator dataset
that was previously analyzed in \cite{cavicchia2022gaussian}, and we acknowledge the authors for providing us with data access.
It contains data on $p=11$ variables related to well-being: education, jobs, income, safety, health, environment, civic engagement, accessibility to services, housing, community, and life satisfaction on which $n=36$ countries are given a score ranging from 0 to 10.

Note, however, that we split the sample of countries into two groups since
\cite{cavicchia2022gaussian} have found evidence for two groups of countries ($n_1=21$ and $n_2=14$) with distinct clustering structures. The first group consists of the countries Australia, Austria, Belgium, Canada, Denmark, Finland, France, Germany, Iceland, Ireland, Italy, Japan, Luxembourg, Netherlands, New Zealand, Norway, Spain, Sweden, Switzerland, United Kingdom, United States, the second group consists of the countries Chile, Czech Republic, Estonia, Greece, Hungary, Israel, Korea, Latvia, Lithuania, Mexico, Poland, Portugal, Slovak Republic, Slovenia, Turkey. In our analysis, we omit Lithuania from the second group due to a missing value, hence $n_1 = 21$ and $n_2 = 14$.

\subsection{Humor Styles Questionnaire} \label{appendix:HSQ}
We analyze data on the humor styles questionnaire (HSQ) developed by \citet{martin2003HSQ}. The $p=32$ items of the HSQ, grouped by the humor style being measured, are listed in Table~\ref{tab:HSQ} together with their position in the survey.
We use responses to the HSQ on a five-point rating scale (anchored by 1 = ``never or very rarely true'' to 5 = ``very often or always true''), which we obtained from \url{https://openpsychometrics.org/_rawdata/}. In addition to the aforementioned $p=32$ items, participants were asked at the end of the questionnaire to indicate the accuracy of their responses. We restrict our analysis to participants who reported that their responses are fully accurate, and after removing 9 observations with missing responses, we retain $n=182$ respondents.

We apply the CGGM algorithm for clustering the covariance matrix (including the
refitting step). It is worth pointing out that the objective function of CGGM is based on the likelihood under the assumption of a normal distribution, which is clearly violated here due to the discrete nature of the data. Nevertheless, it is a common assumption in the behavioral sciences that such rating-scale data are discrete measurements of latent normally-distributed sentiments. Hence, the assumed normal distribution in CGGM may be viewed as a (crude) approximation. Moreover, it is interesting to see whether we can obtain meaningful results even if this normality assumption is violated.

\begin{figure}[!b]
\centering
\includegraphics[width = 0.86\textwidth]{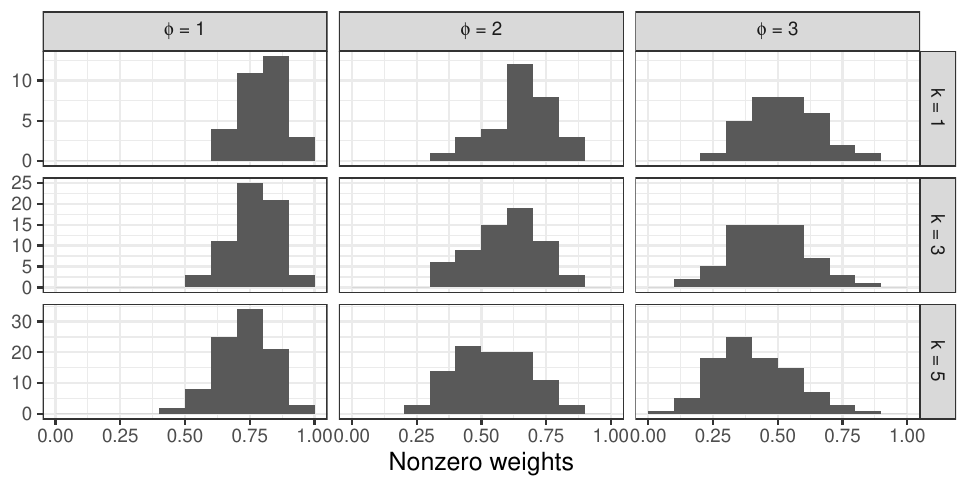}
\caption{Histogram of nonzero clustering weights in CGGM for different candidate values of the tuning parameters $k$ (in rows) and $\phi$ (in columns).}
\label{fig:HSQ_weights}
\end{figure}

Moreover, we use 5-fold cross-validation to determine the optimal values of the tuning parameters $k$, $\phi$, $\lambda_{c}$, and $\lambda_{s}$. We thereby use the same candidate values as in the simulations from Section~\ref{sec:estimate_cov} of the main text. Figure~\ref{fig:HSQ_weights} reveals that the choice of candidate values $k~\in~\{ 1, 3, 5 \}$ and $\phi \in \{ 1, 2, 3\}$ is reasonable due to the amount of variation among the nonzero clustering weights, particularly for $\phi = 3$. In fact, we first simply set $\phi = 3$ after inspecting this plot (only tuning the remaining hyperparameters), and only afterward included $\phi \in \{ 1, 2, 3\}$ in the cross-validation as a robustness check. We obtained the same results in both cases.

\begin{figure}[!h]
\centering
\includegraphics[width = \textwidth]{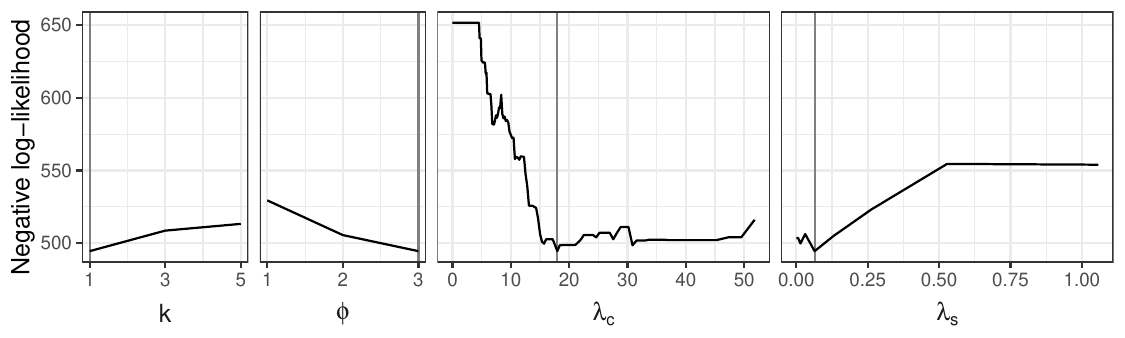}
\caption{Cross-validation score (negative log-likelihood) using the optimal combination of tuning parameters as a baseline. Separate panels visualize the score as one of the tuning parameter varies, with vertical reference lines highlighting the respective optimal values.}
\label{fig:HSQ_cv}
\end{figure}

Finally, we again investigate the stability of the tuning parameter selection by visualizing the cross-validation scores in Figure~\ref{fig:HSQ_cv}. Although the plot for $\lambda_{c}$, and to a lesser extent that of $\lambda_{s}$ show some wiggles, they indicate a clear overall trend through the range of the tuning parameters. Hence, we can be confident that the found optimal values are in a good neighborhood of the tuning parameter space.

\begingroup
\spacingset{1}
\captionsetup{width=\textwidth}
\begin{longtable}{l l@{ } p{11.32cm}}
    \noalign{\smallskip}\toprule\noalign{\smallskip}
    Humor style & \multicolumn{2}{l}{Items} \\
    \noalign{\smallskip}\midrule\noalign{\smallskip}
    \endfirsthead

    \noalign{\smallskip}\toprule\noalign{\smallskip}
    Humor style & \multicolumn{2}{l}{Items} \\
    \noalign{\smallskip}\midrule\noalign{\smallskip}
    \endhead

    \noalign{\smallskip}\bottomrule
    \noalign{\medskip}
    \caption[]{\spacingset{1.75}List of the $p = 32$ items of the humor styles questionnaire of \citet{martin2003HSQ}, grouped by the humor style being measured and reported together with their position in the survey. Items marked with an asterisk are reverse-keyed, i.e., the responses on the five-point rating scale are reversed prior to analysis.}\\
    \multicolumn{3}{r}{\textit{Table continues on the next page}}
    \endfoot

    \noalign{\smallskip}\bottomrule
    \noalign{\medskip}
    \caption[]{\spacingset{1.75}List of the $p = 32$ items of the humor styles questionnaire of \citet{martin2003HSQ}, grouped by the humor style being measured and reported together with their position in the survey. Items marked with an asterisk are reverse-keyed, i.e., the responses on the five-point rating scale are reversed prior to analysis.} \\
    \endlastfoot

Affiliative
    & 1.  & I usually don't laugh or joke around much with other people.$^{*}$ \\
    & 5.  & I don't have to work very hard at making other people laugh---I seem to be a naturally humorous person. \\
    & 9.  & I rarely make other people laugh by telling funny stories about myself.$^{*}$ \\
    & 13. & I laugh and joke a lot with my closest friends. \\
    & 17. & I usually don't like to tell jokes or amuse people.$^{*}$ \\
    & 21. & I enjoy making people laugh. \\
    & 25. & I don't often joke around with my friends.$^{*}$ \\
    & 29. & I usually can't think of witty things to say when I'm with other people.$^{*}$ \\
\noalign{\smallskip}
Self-enhancing
    & 2.  & If I am feeling depressed, I can usually cheer myself up with humor. \\
    & 6.  & Even when I'm by myself, I'm often amused by the absurdities of life. \\
    & 10. & If I am feeling upset or unhappy I usually try to think of something funny about the situation to make myself feel better. \\
    & 14. & My humorous outlook on life keeps me from getting overly upset or depressed about things. \\
    & 18. & If I'm by myself and I'm feeling unhappy, I make an effort to think of something funny to cheer myself up. \\
    & 22. & If I am feeling sad or upset, I usually lose my sense of humor.$^{*}$ \\
    & 26. & It is my experience that thinking about some amusing aspect of a situation is often a very effective way of coping with problems. \\
    & 30. & I don't need to be with other people to feel amused---I can usually find things to laugh about even when I'm by myself. \\
\noalign{\smallskip}
Aggressive
    & 3.  & If someone makes a mistake, I will often tease them about it. \\
    & 7.  & People are never offended or hurt by my sense of humor.$^{*}$ \\
    & 11. & When telling jokes or saying funny things, I am usually not very concerned about how other people are taking it. \\
    & 15. & I do not like it when people use humor as a way of criticizing or putting someone down.$^{*}$ \\
    & 19. & Sometimes I think of something that is so funny that I can't stop myself from saying it, even if it is not appropriate for the situation. \\
    & 23. & I never participate in laughing at others even if all my friends are doing it.$^{*}$ \\
    & 27. & If I don't like someone, I often use humor or teasing to put them down. \\
    & 31. & Even if something is really funny to me, I will not laugh or joke about it if someone will be offended.$^{*}$ \\
\noalign{\smallskip}
Self-defeating
    & 4.  & I let people laugh at me or make fun at my expense more than I should. \\
    & 8.  & I will often get carried away in putting myself down if it makes my family or friends laugh. \\
    & 12. & I often try to make people like or accept me more by saying something funny about my own weaknesses, blunders, or faults. \\
    & 16. & I don't often say funny things to put myself down.$^{*}$ \\
    & 20. & I often go overboard in putting myself down when I am making jokes or trying to be funny. \\
    & 24. & When I am with friends or family, I often seem to be the one that other people make fun of or joke about. \\
    & 28. & If I am having problems or feeling unhappy, I often cover it up by joking around, so that even my closest friends don't know how I really feel. \\
    & 32. & Letting others laugh at me is my way of keeping my friends and family in good spirits.
\label{tab:HSQ}
\end{longtable}
\endgroup

\end{document}